\documentclass{article}
\usepackage[accepted]{icml2018}
\usepackage{microtype}
\usepackage{graphicx}
\usepackage{subfigure}
\usepackage{booktabs} 

\usepackage{rbase}
\usepackage{amsmath}
\usepackage{amsfonts}
\usepackage{amssymb}
\usepackage{amsthm}
\usepackage{gensymb,stmaryrd,mathtools,textcomp,xspace,grffile}

\usepackage{color}
\definecolor{hcitecolor}{RGB}{40,40,160}
\usepackage[colorlinks=true,citecolor=hcitecolor,linkcolor=black,urlcolor=blue!50!black]{hyperref}
\usepackage{tikz}
\usepackage{graphicx}
\usepackage{pgfplots}
\usepackage{caption}
\usepackage{subfigure}
\usepackage{graphics}
\usepackage{xypic}
\usepackage[all,2cell,ps,cmtip]{xy}
\usepackage{tikz}
\usepackage{epsfig}
\usepackage{enumerate}
\usepackage{stmaryrd}
\usepackage{bbm}
\usepackage{algorithm}
\usepackage{algorithmic}
\usepackage{paralist}
\usepackage{soul}
\usepackage{amscd}
\usepackage{tabularx}

\usepackage{hyperref}

\makeatletter
\def\thm@space@setup{%
	\thm@preskip=5pt
	\thm@postskip=5pt 
}
\makeatother

\newtheorem{thm}{Theorem}
\newtheorem{prop}{Proposition}

\newtheorem{lem}{Lemma}

\newtheorem{defn}{Definition}
\newtheorem{exmpl}{Example}

\newenvironment{pf}{\textbf{Proof.}}{\mbox{}\hfill\m{\blacksquare}\\} 
\newenvironment{pfof}[1]{\textbf{Proof of #1.}}{\mbox{}\hfill\m{\blacksquare}\\ \mbox{}\noindent}

\newcommand{\framedef}[1]{\centerline{\fbox{~\parbox{.95\columnwidth}{\vspace{-4pt} #1 \vspace{-4pt}}~}}}

\tikzset{cnode/.style={circle,draw,thick,minimum size=1.25em}}

\tikzset{cnodesmall/.style={circle,draw,thick,minimum size=.5em}}

\usetikzlibrary{backgrounds,arrows}
\pgfdeclarelayer{background}
\pgfsetlayers{background,main}
\usetikzlibrary{decorations}
\usetikzlibrary{shapes}
\usetikzlibrary{topaths}
\usetikzlibrary{calc,3d}

\newenvironment{mequation*}{$}{$}

\newcommand{\iinterval}[2]{\llbracket\tts #1,#2\rrbracket}
\newcommand{\lins}[2]{L_{#2}(#1)}

\newcommand{\Tact}{\mathbb{T}}

\newcommand{\coset}[2]{[\ts #1\ts]_{\scriptstyle #2}}
\newcommand{\gconv}{\ast}
\newcommand{\crepr}[1]{\ensuremath{\wbar{#1}}}

\newcommand{\actfn}{\sigma}

\newcommand{\tupr}[1]{\tup^{\raisebox{-4pt}{\scriptsize\m{#1}}}}

\newcommand{\neuron}{\mathfrak{n}}

\ignore{
\newcommand{\pskip}{\smallskip \vspace{5pt}\\ \noindent}

}

\icmltitlerunning{On the Generalization of Equivariance and Convolution in Neural Networks}

\begin{document}
\twocolumn[
\icmltitle{On the Generalization of Equivariance and Convolution in Neural Networks \\ 
to the Action of Compact Groups}




\begin{icmlauthorlist}
\icmlauthor{Risi Kondor}{to}
\icmlauthor{Shubhendu Trivedi}{goo}
\end{icmlauthorlist}

\icmlaffiliation{to}{Departments of Statistics and Computer Science, The University of Chicago}
\icmlaffiliation{goo}{Toyota Technological Institute at Chicago}

\icmlcorrespondingauthor{Risi Kondor}{\href{mailto:risi@cs.uchicago.edu}{risi@cs.uchicago.edu}}
\icmlcorrespondingauthor{Shubhendu Trivedi}{\href{mailto:shubhendu@ttic.edu}{shubhendu@ttic.edu}}

\icmlkeywords{Machine Learning, ICML}

\vskip 0.3in
]



\printAffiliationsAndNotice{}  

\begin{abstract}
Convolutional neural networks have been extremely successful in the image recognition domain 
because they ensure equivariance to translations. 
There have been many recent attempts to generalize this framework to other domains, including 
graphs and data lying on manifolds. 
In this paper we give a rigorous, theoretical treatment of convolution and equivariance in neural 
networks with respect to not just translations, but the action of any compact group. 
Our main result is to prove that (given some natural constraints) convolutional structure is 
not just a sufficient, but also a necessary condition for equivariance to the action of a compact group. 
Our exposition makes use of concepts from representation theory and noncommutative harmonic analysis 
and derives new generalized convolution formulae.

\ignore{
One of the prime reasons for the success of convolutional neural networks in the image recognition
domain is that they are equivariant to translations of the input, providing a useful inductive bias
for the task. More generally, constructing architectures that are equivariant to symmetry
transformations is a rational and principled approach to neural network design. Unsurprisingly,
there has been significant recent work in this space: researchers have designed networks operating
on images that are equivariant to not just translations, but also rotations, wraps and flips. There
has also been activity on constructing neural networks for specific domains (such as on
$\mathcal{S}^2$), by deriving the right notion of convolution and showing that the network is
equivariant to the desired symmetry transformations. However, it is not immediately obvious how to
generalize the notion of convolution to general domains and combinatorial objects such as
graphs. Moreover, the relationship between equivariance and convolution in the general case is
unclear. In this paper, we bridge this gap in our understanding by achieving the following two
goals: (a) We prove that convolution is a necessary and sufficient condition for equivariance in
feed-forward networks, not just in image recognition, but in the general context of
convolution/equivariance w.r.t.\;the action of \emph{any} compact group on a sequence of homogeneous
spaces. (b) We argue that in many cases convolution alone is not rich enough to capture structure,
so we introduce a generalization called steerable convolution, and again prove that it is necessary
and sufficient for the corresponding notion of equivariance.
}
	
\end{abstract}

\section{Introduction}

One of the most successful neural network architectures is convolutional neural networks (CNNs) 
\citep{LeCun1989}. 
In the image recognition domain, where CNNs were originally conceived, convolution plays two crucial roles.   
First, it ensures that in any given layer, exactly the same filters are applied to each part of the image. 
Consequently, if the input image is translated, the activations of the network in each layer will translate the 
same way. This property is called \emph{equivariance} \citep{Cohen2016}. 
Second, in conjunction with pooling, convolution ensures that each neuron's effective receptive field 
is a spatially contiguous domain.
As we move higher in the network, these domains generally get larger,  
allowing the CNN to capture structure in images at \emph{multiple different scales}.

Recently, there has been considerable interest in extending neural networks to more exotic types 
of data, such as graphs or functions on manifolds  
\citep{Niepert2016,Defferrard2016,Duvenaud2015,Li2016,SphericalCNN2018, Monti2017Geometric, Masci2015Geodesic}.  
In these domains, equivariance and multiscale structure are just as 
important as for images, but finding the right notion of convolution 
is not obvious. 

On the other hand, mathematics does offer a sweeping generalization of convolution tied in deeply  
with some fundamental ideas of abstract algebra:   
if \m{G} is a compact group  
and \m{f} and \m{g} are two functions \m{G\to\CC}, then the convolution of \m{f} with \m{g} is defined 
\begin{equation}\label{eq: convo0}
(f\ast g)(u)=\int_G f(uv^{-1})\,g(v)\,d\mu(v).
\end{equation}
Note the striking similarity of this formula to the ordinary notion of convolution, except that 
in the argument of \m{f}, \m{u-v} has been replaced by the group operation \m{u v^{-1}}, 
and integration is with respect to the Haar measure, \m{\mu}. 

The goal of this paper is to relate \rf{eq: convo0} to the various looser notions of convolution 
used in the neural networks literature, and show that several practical neural networks implicitly 
already take advantange of the above group theoretic concept of convolution. 
In particular, we prove the following theorem (paraphrased here for simplicity).

\begin{thm}
A feed forward neural network \m{\Ncal} is equivariant to the action of a compact group \m{G} on its inputs 
if and only if each layer of \m{\Ncal} implements a generalized form of convolution derived from \rf{eq: convo0}. 
\end{thm}
\setcounter{thm}{0}

To the best of our knowledge, this is the first time that the connection 
between equivariance and convolution in neural networks has been stated at this level of generality. 
One of the technical challenges in proving our theorem is that 
the activations in each layer of a neural net 
correspond to functions on a sequence of spaces \emph{acted on} by \m{G} (called \emph{homogeneous spaces} 
or \emph{quotient spaces}) rather than functions on \m{G} itself. 
This necessitates a discussion of group convolution that is rather more thoroughgoing than is 
customary in pure algebra.

This paper does not present any new algorithms or neural network architectures. 
Rather, its goal is to provide the language for thinking about generalized notions of equivariance and 
convolution in neural networks, and thereby facilitate the development of future architectures 
for data with non-trivial symmetries.  
To avoid interruptions in the flow of our exposition, we first present the theory in its abstract 
form, and then illustrate it with examples in Section \ref{sec: examples}. 
For better understanding, the reader might choose to skip back and forth between these sections. 
One work that is close in spirit to the present paper but only considers discrete groups is 
\citep{Poczos2017}.  
\section{Notation}\label{sec: notation}

In the following \m{[a]} will denote the set \m{\cbrN{\oneton{a}}}. 
Given a set \m{\Xcal} and a vector space \m{V},\:  
\m{\lins{\Xcal}{V}} will denote the space of functions \m{\cbrN{f\colon \Xcal\to V}}.

\ignore{
\begin{table*}[t]
\begin{tabularx}{\textwidth}{|p{70pt}|X|}
\hline
\m{\iinterval{a}{b}}&The set \m{\cbrN{a,a\<+1,\ldots,b}}.\\ 
\m{\iota}&The imaginary unit, \m{\iota=\sqrt{-1}}.\\
\m{M_{i,j}} or \m{[M]_{i,j}}&The \m{(i,j)} element of the matrix \m{M}.\\
\m{[M]_{i,\ast},\:[M]_{\ast,j}}&The \m{i}'th row and \m{j}'th column of \m{M}, respectively.\\
\m{M^\dag}&The Hermitian conjugate (conjugate transpose) of \m{M}.\\
\m{\lins{\Xcal}{V}}&The space of functions \m{\cbrN{f\colon \Xcal\to V}}\\
\hline
\m{\Xcal_\ell}&The index set of the \m{\ell}'th layer in a feed-forward network (page \pageref{p: Xcal}).\\
\m{f_\ell}&The representation formed by the \m{\ell}'th layer of the network (page \pageref{p: fell}).\\
\m{\phi_\ell}&The linear map realized by the \m{\ell}'th layer (Definition \ref{def: NN}).\\ 
\m{\actfn_\ell}&The nonlinearity at layer \m{\ell} (Definition \ref{def: NN}).\\ 
\hline
\m{G}&An abstract group, in our case almost always compact (Appendix A).\\
\m{\cbrN{T_g}_{g\in G}}&The action \m{\cbrN{T_g\colon S\to S}_{g\in G}} of \m{G} on some set \m{S} (Definition \ref{def: action}).\\
\m{\cbrN{\Tact_g}_{g\in G}}&The translation action of \m{G} on a homogeneous space (Definition \ref{def: translation}).\\
\hline
\m{H\leq G}& \m{H} is a subgroup of \m{G} (Appendix A).\\
\m{gH,\:Hg,\:KgH}&The left, right and double cosets of a group element \m{g} (page \pageref{p: coset}).\\
\m{G/H,\:H\backslash G}&The left and right quotient spaces of \m{G} w.r.t.\:\m{H} (page \pageref{p: qspace}).\\
\m{K\backslash G/H}&The double quotient space of \m{G} w.r.t.\:\m{H} and \m{K} (page \pageref{p: qspace}).\\
\m{\coset{g}{\Xcal}}&The point \m{g(x_0)\tin\Xcal} corresponding to \m{g\tin G} (page \pageref{p: cosetn}).\\
\m{\crepr{x}}&The representative of the coset that maps \m{x_0} to \m{x} (page \pageref{p: cosetn}).\\
\m{f\tdown_{G/H}}&The projection of \m{f} to \m{G/H} (page \pageref{p: proj}).\\ 
\m{f\tup_{G}}&The lifting of \m{f} to \m{G}, also called induced function (page \pageref{p: proj}).\\ 
\hline
\m{f\ast g}&The convolution of \m{f} with \m{g} (Definitions \ref{def: convo gp}--\ref{def: convo homov}).\\
\m{f\star g}&The cross-correlation of \m{f} with \m{g} (page \ref{p: corr}).\\
\m{\inp{f,g}}&The inner product \m{\sum_{u\in G}f(u)^\ast g(u)}\\
\m{\h f}&The Fourier transform of \m{f} (pages \pageref{p: fourier-beg}--\pageref{p: fourier-end}).\\
\hline
\m{\rho}&A specific representations of a group (Appendix B).\\
\m{d_\rho}&The dimensionality of the representation \m{\rho} (Appendix B).\\
\m{\rho\vert_H}&The restriction of \m{\rho} to a subgroup \m{H} (Appendix B).\\
\m{\Rcal_G}&A complete set of inequivalent irreducible representations of \m{G}.\\
\hline
\end{tabularx}
\caption{\label{tbl: notation}}
\end{table*}
}
\section{Equivariance in neural networks}

A feed-forward neural network consists of some number of  
``neurons'' arranged in \m{L\<+1} distinct layers. 
Layer \m{\ell\<=0} is the input layer, where data is presented to the network, while layer 
\m{\ell\<=L} is where the output is read out. 
Each neuron \m{\neuron^\ell_x} (denoting neuron number \m{x} in layer \m{\ell}) has an \emph{activation} 
\m{f^\ell_x}. For the input layer, the activations come directly from the data, whereas  
in higher layers they are computed via a simple function of the activations of the previous layer, such as 
\vspace{-3pt}
\begin{equation}\label{eq: neuron1}
f^\ell_x=\actfn\brbig{b^\ell_x+\textstyle \sum_{y} w^\ell_{x,y}\,f^{\ell-1}_y}.
\end{equation}
Here, the \m{\cbrN{b^\ell_x}} bias terms and the \m{\cbrN{w^{\ell}_{x,y}}} weights are the network's 
learnable parameters, while \m{\actfn} is a fixed nonlinear function, such as 
the ReLU function \m{\actfn(z)\<=\max(0,z)}. 
In the simplest case, each \m{f^\ell_x} is a scalar, but, 
in the second half of the paper we consider neural networks with more general,  
vector or tensor valued activations.

For the purposes of the following discussion it is actually helpful to take a slightly more abstract 
view, and, instead of focusing on the individual activations, consider the activations in any given layer 
collectively as a function \m{f^\ell\colon \Xcal_\ell\<\to V_\ell}, 
where \m{\Xcal_\ell} is a set indexing the neurons and \m{V_\ell} is a vector space.  
Omitting the bias terms in \rf{eq: neuron1} for simplicity, each layer \m{\ell=1,2,\ldots,L} can then 
just be thought of as implementing a linear transformation 
\m{\phi_\ell\colon \lins{\Xcal_{\ell-1}}{V_{\ell-1}} \to \lins{\Xcal_{\ell}}{V_\ell}} followed by 
the pointwise nonlinearity \m{\sigma}. 
Our operational definition of neural networks for the rest of this paper will be as follows. 

\begin{defn}\label{def: NN}
Let \m{\sseqz{\Xcal}{L}} be a sequence of index sets, \m{\sseqz{V}{L}} vector spaces, 
\m{\sseq{\phi}{L}} linear maps\vspace{-5pt} 
\[\phi_\ell\colon \lins{\Xcal_{\ell-1}}{V_{\ell-1}} \longrightarrow \lins{\Xcal_{\ell}}{V_\ell},
\vspace{-5pt}
\]
and \m{\actfn_\ell\colon V_\ell\to V_\ell} appropriate pointwise nonlinearities, such as the ReLU operator. 
The corresponding \df{multilayer feed-forward neural network (MFF-NN)} is then a sequence of 
maps \m{f_0\mapsto f_1\mapsto f_2\mapsto \ldots\mapsto f_L}, where\;  
\begin{mequation*}
f_\ell(x)=\actfn_\ell(\phi_\ell(f_{\ell-1})(x)).
\end{mequation*}
\end{defn}

If we are interested in constructing a neural net for recognizing \m{m\times m} pixel images, it is tempting 
to take \m{\Xcal_0=[m]\times [m]} and define \m{\sseq{\Xcal}{L}} similarly. However, again for notational 
simplicity, we extend each of these index sets to the entire integer plane \m{\ZZ^2}, and simply assume 
that outside of the square region \m{[m]\times [m]}, \m{f^0(x_1,x_2)\<=0}. 
A traditional \emph{convolutional neural network (CNN)} is a network of this type where the \m{\phi_\ell} 
functions are constrained to have the special form \vspace{-5pt}
\begin{multline}\label{eq: convo1}
\phi_\ell(f_{\ell-1})(x_1,x_2)=\\
\sum_{u_1=1}^w \sum_{u_2=1}^w  f_{\ell-1}(x_1\<-u_1,x_2\<-u_2)\;\chi_\ell(u_1,u_2). 
\vspace{-2pt}
\end{multline}
The above function is known as the \emph{discrete convolution} of \m{f^{\ell-1}} with the \emph{filter} 
\m{\chi}, and is usually denoted \m{f_{\ell-1}\ast \chi_\ell}. 
In most CNNs the width \m{w} of the filters is quite small, on the order of \m{3\sim 10}, while 
the number of layers can be as small as 3 or as large as a few dozen. 

Some of the key features of CNNs are immediately apparent from the convolution formula \rf{eq: convo1}:
\begin{compactenum}[~1.]
\item The number of parameters in CNNs is much smaller than in general (fully connected) feed-forward 
networks, since we only have 
to learn the \m{w^2} numbers defining the \m{\chi_\ell} filters rather than \m{O((m^2)^2)} weights. 
\item \rf{eq: convo1} applies the same filter to every part of the image. Therefore, if the networks 
learns to recognize a certain feature, e.g., eyes, in one part of the image, then it will be 
able to do so in any other part as well. 
\item Equivalently to the above, if the input image is translated by any vector 
\m{(t_1,t_2)} (i.e., \m{{f^0}'(x_1,x_2)=f^0(x_1\<-t_1,x_2\<-t_2)}, then all higher layers 
will translate in exactly the same way. This property is called \df{equivariance} 
(sometimes \emph{covariance}) to translations.   
\end{compactenum}  
The goal of the present paper is to understand the mathematical generalization of the above properties to 
other domains, such as graphs, manifolds, and so on.

\subsection{Group actions}

The jumping off point to our analysis is the observation that the above is a special case of the 
following scenario. 
\begin{compactenum}[~1.]
\item We have a set \m{\Xcal} and a function \m{f\colon\Xcal\to\CC}. 
\item We have a group \m{G} acting on \m{\Xcal}. This means that each \m{g\tin G} has a 
corresponding transformation \m{T_g\colon\Xcal\to\Xcal},  
and for any \m{g_1,g_2\tin G},\; \m{T_{g_2 g_1}\<=T_{g_2}\circ T_{g_1}}.
\item The action of \m{G} on \m{\Xcal} extends to functions on \m{\Xcal} by \vspace{-5pt} 
\[\Tact_g\colon f\mapsto f'\qqquad f'(T_g(x))=f(x).\] 
\end{compactenum}
\smallskip

\noindent 
In the case of translation invariant image recognition, \m{\Xcal\<=\ZZ^2}, \m{G} is the group of 
integer translations, which is isomorphic to \m{\ZZ^2} (note that this is a very special case, in general 
\m{\Xcal} and \m{G} are different objects), the action is \vspace{-3pt}
\[T_{(t_1,t_2)}(x_1,x_2)=(x_1\<+t_1,x_2\<+t_2)\qquad (t_1,t_2)\tin\ZZ^2,\]
and the corresponding (induced) action on functions is 
\[\Tact\colon f\mapsto f'\qquad f'(x_1,x_2)=f(x_1\<-t_1,x_2\<-t_2).\]  
We give several other (more interesting) examples of group actions in Section \ref{sec: examples}, 
but for now continue with our abstract development. 
Also note that to simplify notation, in the following, where this does not cause confusion,  
we will simply write group actions as \m{x\mapsto g(x)} rather than the more cumbersome \m{x\mapsto T_g(x)}. 

Most of the actions considered in this paper have the property that taking any \m{x_0\tin \Xcal}, 
any other \m{x\tin\Xcal} can be reached by the action of some \m{g\tin G}, i.e., \m{x\<=g(x_0)}. 
This property is called \df{transitivity}, and if the action of \m{G} on \m{\Xcal} is transitive, 
we say that \m{\Xcal} is a \df{homogeneous space} of \m{G}.

\subsection{Equivariance}

Equivariance is a concept that applies very broadly, whenever we have a group acting on 
a pair of spaces and there is a map from functions on one to functions on the other. 
\begin{defn}\label{def: equivariance}
Let \m{G} be a group and \m{\Xcal_1,\Xcal_2} be two sets with corresponding \m{G}-actions \vspace{-3pt}
\[T_g\colon \Xcal_1\to \Xcal_1,\qqquad\qqquad T_g'\colon \Xcal_2\to \Xcal_2. \vspace{-3pt}
\]
Let \m{V_1} and \m{V_2} be vector spaces, and \m{\Tact} and \m{\Tact'} be the induced actions of 
\m{G} on \m{\lins{\Xcal_1}{V_1}} and \m{\lins{\Xcal_2}{V_2}}. 
We say that a (linear or non-linear) map \m{\phi\colon \lins{\Xcal_1}{V_1}\to \lins{\Xcal_2}{V_2}} is 
\df{equivariant} with the action of \m{G} 
(or \m{G}\df{--equivariant} for short) if \vspace{-3pt}
\[\phi(\Tact_g(f))=\Tact'_g(\phi(f))\qqquad \forall f\tin \lins{\Xcal_1}{V_1} \vspace{-3pt}
\]
for any group element \m{g\tin G}.   
\end{defn}

Equivariance is represented graphically by a so-called commutative diagram, in our case 
\ignore{
\[\begin{CD}
S_1 @>{T_g}>> S_1\\
@V{\phi}VV @V{\phi}VV \vspace{-5pt}\\ 
S_2 @>{T'_g}>> S_2\\ 
\end{CD}\]
}
\[\xymatrix@R=30pt@C=30pt{
\lins{\Xcal_1}{V_1} \ar[r]^{\Tact_g} \ar[d]^{\phi}& \lins{\Xcal_1}{V_1}\ar[d]^{\phi}\\ 
\lins{\Xcal_2}{V_2} \ar[r]^{\Tact'_g}& \lins{\Xcal_2}{V_2}\\ 
}\]
We are finally in a position to define the objects that we study in this paper, namely 
generalized equivariant neural networks. 
\bigskip

\framedef{
\begin{defn}\label{def: equivariant NN}
Let \m{\Ncal} be a feed-forward neural network as defined in Definition \ref{def: NN},  
and \m{G} be a group that acts on each index space \m{\sseqz{\Xcal}{L}}.  
Let \m{\Tact^0,\Tact^1,\ldots,\Tact^L} be the corresponding actions 
on \m{\lins{\Xcal_{0}}{V_{0}},\ldots,\lins{\Xcal_{L}}{V_{L}}}. 
We say that \m{\Ncal} is a \m{G}--\df{equivariant feed-forward network} if,   
when the inputs are transformed \m{f_0\mapsto \Tact^{0}_g(f_0)} (for any \m{g\tin G}), 
the activations of the   
other layers correspondingly transform as \m{f_\ell\mapsto \Tact^\ell_g(f_\ell)}. 
\end{defn}
}
\smallskip

\noindent 
It is important to note how general the above framework is. In particular, we have not said whether 
\m{G} and \m{\sseqz{\Xcal}{L}} are discrete or continuous. 
In any actual implementation of a neural network, the index sets would of course be finite. 
However, it has been observed before that in certain cases, specifically when \m{\Xcal_0} is 
an object such as the sphere or other manifold which does not have a discretization that fully takes 
into account its symmetries, it is easier to describe the situation in terms of abstract 
``continuous'' neural networks than seemingly simpler discrete ones \citep{SphericalCNN2018}.  

Note also that invariance is a special case of equivariance, where \m{T_g\<=\textrm{id}} for all \m{g}. 
In fact, this is another major reason why equivariant architectures are so prevalent in the 
literature: any equivariant network can be turned into a \m{G}--invariant network 
simply by tacking on an extra layer that is equivariant in this 
degenerate sense (in practice, this often means either averaging or creating a histogram of 
the activations of the last layer). 
Nowhere is this more important than in graph learning, where it is a hard constraint that 
whatever representation is learnt by a neural network, it must be invariant to reordering 
the vertices. Today's state of the art solution to this problem are message passing networks 
\citep{Riley2017}, whose invariance behavior we discuss in section \ref{sec: examples}. 
Another architecture that achieves invariance by stacking equivariant layers followed 
by a final invariant one is that of  scattering networks \citep{MallatScattering}. 

\section{Convolution on groups and quotient spaces}\label{sec: convo}\label{p: fourier-beg}

According to its usual definition in signal processing, 
the \df{convolution} of two functions \m{f,g\colon\RR\to\RR} is 
\begin{equation}\label{eq: convolution}
(f\ast g)(x)=\int f(x\<-y)\,g(y)\,dy.
\end{equation}
Intuitively, we can think of \m{f} as a template and \m{g} as a modulating function 
(or the other way round, since convolution on \m{\RR} is commutative): 
we get \m{f\ast g} by a placing a ``copy'' of \m{f} at each point on the \m{x} axis, but scaled 
by the value of \m{g} at that point, and superimposing the results. 
The discrete variant of \rf{eq: convolution} for \m{f,g\colon \ZZ\to\RR} is of course
\begin{equation}\label{eq: conv discrete}
(f\ast g)(x)=\sum_{y\in\ZZ}f(x-y)\,g(y), 
\end{equation}
and both the above formulae have natural generalizations to higher dimensions. 
In particular, \rf{eq: convo1} is just the two dimensional version of \rf{eq: conv discrete} 
with a limited width filter.

What we are interested in for this paper, however, is the much broader generalization 
of convolution to the case when \m{f} and \m{g} are functions on a compact group \m{G}.  
As mentioned in the Introduction, this takes the form  
\begin{equation}\label{eq: convo gp0}
(f\ast g)(u)=\int_G f(u v^{-1})\,g(v)\:d\mu(v).  
\end{equation} 
Note that \rf{eq: convo gp0} only differs from \rf{eq: convolution} in that \m{x\<-y} is replaced by 
the group operation \m{u v^{-1}}, which is not surprising, since the group operation on \m{\RR} in fact 
is exactly \m{(x,y)\mapsto x\<+y}, and the ``inverse'' of \m{y} in the group sense is \m{-y}.  
Furthermore, the Haar measure \m{\mu} makes an appearance. At this point, the main 
reason that we restrict ourselves to compact groups is because this guarantees that \m{\mu} is 
essentially unique\footnote{Non-compact groups would also cause trouble because their representation 
theory is much more involved. 
\m{\RR^2}, which is the group behind traditional CCNs, is of course not compact. 
The reason that it is still amenable to our analysis (with small modifications) 
is that it belongs to one of a handful of families 
of exceptional non-compact groups that are easy to handle.}. 
The discrete counterpart of \rf{eq: convo gp0} for countable (including finite) groups is 
\begin{equation}\label{eq: convo gp1}
(f\ast g)(u)=\sum_{v\in G} f(uv^{-1})\,g(v).
\end{equation}
All these definitions are standard and have deep connections to the algebraic properties of groups. 
In contrast, the various extensions of convolution to homogeneous spaces that we derive below are not 
often discussed in pure algebra. 

\begin{figure}[t]
\centerline{\fbox{
\begin{minipage}{.95\columnwidth}
\textsc{Essential definitions for quotient spaces}\\ \vspace{-10pt}\\ \noindent  
Certain connections between the structure of a group \m{G} and its homogeneous space \m{\Xcal} 
are crucial for our exposition. First, by definition, fixing an ``origin'' \m{x_0\tin\Xcal}, 
any \m{x\tin\Xcal} can be reached as \m{x\<=g(x_0)} for some \m{g\tin G}. 
This allows us to ``index'' elements of \m{\Xcal} by elements of \m{G}. 
Since we use this mechanism so often, \ig{in our paper, }we introduce the shorthand 
\m{\coset{g}{\Xcal}\<=g(x_0)}, 
which hides the dependence on the (arbitrary) choice of \m{x_0}.\\
Second, elementary group theory tells us that the set of group elements that fix \m{x_0} 
actually form a subgroup \m{H}. 
By further elementary results (see Appendix), the set of group elements that map \m{x_0\mapsto x} 
is a so-called \df{left coset} \m{gH:=\setofN{gh}{h\tin H}}. The set of all such cosets forms 
the \df{(left) quotient space} \m{G/H}. Therefore, \m{\Xcal} can be identified with \m{G/H}.\\
Now for each \m{gH} coset we may pick a \df{coset representative} \m{g'\tin gH}, and let \m{\crepr{x}} 
denote the representative of the coset of group elements that map \m{x_0} to \m{x}. 
Note that while the map \m{g\mapsto\coset{g}{G/H}} is well defined, the map \m{x\mapsto\crepr{x}} 
going in the opposite direction is more arbitary, 
since it depends on the choice of coset representatives.  
\\
The \df{right quotient space} \m{H\backslash G} is similarly defined as the space of \df{right cosets} 
\m{Hg:=\setofN{hg}{h\tin H}}. Furthermore, if \m{K} is another subgroup of \m{G}, we can talk about 
\df{double cosets} \m{HgK=\setofN{hgk}{h\tin H, k\tin K}} and the corresponding space \m{H\backslash G/K}.   
\\ 
Given \m{f\<\colon G\<\to\CC}, we define its \df{projection} 
to \m{\Xcal\<=G/H}
\vspace{-6pt}
\begin{equation*}\label{eq: proj}
f\tdown_{\Xcal}\colon\Xcal\to\CC \qqquad 
f\tdown_{\Xcal}(x)=\ovr{\abs{H}}\: \sum_{g\in\crepr{x}H} f(g).\vspace{-4pt}
\end{equation*}
Conversely, given \m{f\colon \Xcal\to\CC}, we define the \df{lifting} of \m{f} to \m{G} 
\vspace{-14pt}
\begin{equation*}\label{eq: lifting}
f\tup^G\colon G\to\CC \qqquad 
f\tup^G(g)= f(\coset{g}{\Xcal}).\vspace{-3pt}  
\end{equation*}
Projection and lifting to/from right quotient spaces and double quotient spaces is defined analogously.  
\end{minipage}
\vspace{-8pt}~
}}
\end{figure}

\subsection{Convolution on quotient spaces}

The major complication in neural networks is that \m{\Xcal_0,\ldots,\Xcal_L} 
(which are the spaces that the \m{\sseqz{f}{L}} activations are defined on) are homogeneous spaces of 
\m{G}, rather than being \m{G} itself. 
Fortunately, the strong connection between the structure of groups and their homogeneous spaces (see boxed text)  
allows generalizing convolution to this case as well. 
Note that from now on, to keep the exposition as simple as possible, we present our results assuming that 
\m{G} is countable (or finite). The generalization to continuous groups is straightforward. 
We also allow all our functions to be complex valued, because representation theory itself, which 
is the workhorse behind our results, is easiest to formulate over \m{\CC}. 

\begin{defn}\label{def: convo homo0}
Let \m{G} be a finite or countable group, \m{\Xcal} and \m{\Ycal} be (left or right) quotient spaces of \m{G}, 
\m{f\colon \Xcal\to\CC}, and \m{g\colon \Ycal\to\CC}. 
We then define the \df{convolution} of \m{f} with \m{g} as 
\begin{equation}\label{eq: convo homo0}
(f\ast g)(u)=\sum_{v\in G} f\tup^G(uv^{-1})\:g\tup^G(v), \qquad u\tin G.
\vspace{-5pt}
\end{equation} 
This definition includes \m{\Xcal\<=G} or \m{\Ycal\<=G} as special cases, since any group 
is a quotient space of itself with respect to the trivial subgroup \m{H\<=\cbrN{e}}. 
\end{defn}

\noindent 
Definition \ref{def: convo homo0} hides the facts that depending on the choice of \m{\Xcal} and \m{\Ycal}:  
(a) the summation might only have to extend over a quotient space of \m{G} rather than the entire group, 
(b) the result \m{f\ast g} might have symmetries that effectively make it a function on a quotient 
space rather than \m{G} itself (this is exactly what the case will be in generalized convolutional networks). 
Therefore we now discuss three special cases. 

\subsubsection*{\textbf{Case I:~~ \m{\Xcal\<=G} ~and~ \m{\Ycal\<=G/H}}}

\noindent 
When \m{f\colon G\<\to\CC} but \m{g\colon G/H\<\to\CC} for some subgroup \m{H} of \m{G}, 
\rf{eq: convo homo0} reduces to 
\vspace{-5pt}
\begin{equation*}\label{eq: ch1}
(f\ast g)(u)=\sum_{v\in G} f(uv^{-1})\,g\tup^G(v).\vspace{-5pt}
\end{equation*}
Plugging \m{u'\<=uh} into this formula (for any \m{h\tin H}) and 
changing the variable of summation to \m{w\<{:=}vh^{-1}} gives   
\begin{align*}\label{eq: ch2}
(f\ast g)(u')
&=\sum_{v\in G} f(uhv^{-1})\,g\tup^G(v)\\
&=\sum_{w\in G} f(uw^{-1})\,g\tup^G(wh).\vspace{-5pt}
\end{align*}
However, since \m{w} and \m{wh} are in the same left \m{H}--coset, \m{g\tup^G(wh)=g\tup^G(w)}, 
so \m{(f\ast g)(u')=(f\ast g)(u)}, i.e.,  
\m{f\ast g} is constant on left \m{H}--cosets.  
This makes it natural to interpret \m{f\ast g} as a function on \m{G/H} rather than the full group. 
Thus, we have the following definition. 
\bigskip 

\framedef{
\vspace{5pt}
If \m{f\colon G\<\to\CC}, and \m{g\colon G/H\<\to\CC} then \m{f\<\ast g\colon G/H\to\CC} with  
\vspace{-10pt}
\begin{equation}\label{eq: convo homo1}
(f\ast g)(x)=\sum_{v\in G} f(\crepr{x}v^{-1})\:g(\coset{v}{G/H}). 
\vspace{-5pt}
\end{equation} 
}

\ignore{
\begin{defn}\label{def: convo homo1}
Let \m{G} be a finite or countable group, \m{H} a subgroup of \m{G}, 
\m{f\colon G\to\CC}, and \m{g\colon G/H\to\CC}. 
We then define the \df{convolution} of \m{f} with \m{g} as 
\begin{equation}\label{eq: convo homo1}
f\ast g\colon G/H\to\CC\qquad
(f\ast g)(x)=\sum_{v\in G} f(\crepr{x}v^{-1})\:g(\coset{v}{G/H}),
\vspace{-5pt}
\end{equation} 
where \m{\crepr{x}} denotes the representative of the left \m{H}--coset corresponding to 
\m{x\tin G/H}. 
\end{defn}
}
\bigskip

\subsubsection*{\textbf{Case II:~ \m{\Xcal\<=G/H} ~and~ \m{\Ycal\<=H\backslash G}}}

\noindent 
When \m{f\colon G/H\to\CC}, but \m{g\colon G\to\CC}, \rf{eq: convo homo0} reduces to 
\begin{equation}\label{eq: ch3}
(f\ast g)(u)=\sum_{v\in G} f\tup^G(uv^{-1})\,g(v).
\end{equation}
This time it is not \m{f\<\ast g}, but \m{g} that shows a spurious symmetry. 
Letting \m{v'\<=h v} (for any \m{{h\tin H}}), by the right \m{H}--invariance of 
\m{f\tup^G}, \m{f\tup^G(uv'{}^{-1})=f\tup^G(uv^{-1}{h^{-1})=f\tup^G(uv)}}. 
Considering that any \m{v} can be uniquely written as \m{v=h\crepr{y}}, where \m{\crepr{y}} is the representative 
of one of its cosets, while \m{h\tin H}, we get that \rf{eq: ch3} factorizes in the form 
\begin{align*}\label{eq: ch3b}
(f\ast g)(u)
&=\sum_{y\in H\backslash G} f\tup^G(u\crepr{y}^{-1}) \sum_{h\in H} g(h\crepr{y})\\
&=\sum_{y\in H\backslash G} f\tup^G(u\crepr{y}^{-1})\, \tilde g(y),   
\end{align*}
where \m{\tilde g(y):=\sum_{h\in H} g(h\crepr{y})}. 
In other words, 
without loss of generality we can take \m{g} 
to be a function on \m{H\backslash G} rather than the full group. 
\bigskip 

\framedef{\vspace{5pt}
If \m{f\colon G/H\to\CC}, and \m{g\colon H\backslash G\to\CC}, then  
\m{f\ast g\colon G\to\CC} with\vspace{-4pt} 
\begin{equation}\label{eq: convo homo2}
(f\ast g)(u)=\abs{H} \sum_{y\in H\backslash G} f(\coset{u\crepr{y}^{-1}}{G/H})\,g(y).
\vspace{-5pt}
\end{equation} 
}
\bigskip 

\ignore{
\begin{defn}\label{def: convo homo2}
Let \m{G} be a finite or countable group, \m{H} a subgroup of \m{G},\:  
\m{f\colon G/H\to\CC}, and \m{g\colon H\backslash G\to\CC}. 
We then define the \df{convolution} of \m{f} with \m{g} as \m{f\ast g\colon G\to\CC}, 
\begin{equation}\label{eq: convo homo2}
(f\ast g)(u)=\abs{H} \sum_{y\in H\backslash G} f(\coset{u\crepr{y}^{-1}}{G/H})\,g(y).
\end{equation} 
Note that the constant factor \m{\abs{H}} appears in this equation solely for compatibility with 
\rf{eq: convo homo0}. If \m{H} is infinite, it may be dropped.   
\end{defn}
}

\subsubsection*{\textbf{Case III:~ \m{\Xcal\<=G/H} ~and~ \m{\Ycal\<=H\backslash G/K}}}

\noindent
Finally, we consider the case when \m{f\colon G/H\to\CC} and \m{g\colon G/K\to\CC} for two subgroups 
\m{H,K} of \m{G}, which might or might not be the same. 
This combines features of the above two cases in the sense that, similarly to Case I, 
setting \m{u'\<=uk} for any \m{k\tin K} and letting \m{w=v k^{-1}}, 
\begin{multline*}
(f\ast g)(u')
=\sum_{v\in G} f\tup^G(u'v^{-1})\,g\tup^G(v)=\\
=\sum_{v\in G} f\tup^G(ukv^{-1})\,g\tup^G(v)
=\sum_{w\in G} f\tup^G(uw^{-1})\,g\tup^G(wk)\\
=\sum_{w\in G} f\tup^G(uw^{-1})\,g\tup^G(w)
=(f\ast g)(u),
\end{multline*}
showing that \m{f\ast g} is right \m{K}--invariant, and therefore can be regarded as a function \m{G/K\to\CC}. 
At the same time, similarly to \rf{eq: ch3}, letting \m{v=h\crepr{y}}, 
\begin{align*}
(f\ast g)(u)
&=\sum_{y\in H\backslash G} f\tup^G(u\crepr{y}^{-1}) \sum_{h\in H} g\tup^G(h\crepr{y})\\
&=\sum_{y\in H\backslash G} f\tup^G(u\crepr{y}^{-1})\, \tilde g(y),   
\end{align*}
where \m{\tilde g(y):=\sum_{h\in H} g(h\crepr{y})}, which is left \m{H}--invariant. 
Therefore, without loss of generality, we can take \m{g} to be a function \m{H\backslash G/K\to\CC}. 
\bigskip

\framedef{
\vspace{5pt}
If \m{f\colon G/H\<\to\CC}, and \m{g\colon H\backslash G/K\<\to\CC} 
then we define the \df{convolution} of \m{f} with \m{g} as 
\m{f\ast g\colon G/K\to\CC} with  
\begin{equation}\label{eq: convo homo3}
(f\ast g)(x)=\abs{H} \sum_{y\in H\backslash G} f(\coset{\crepr{x} \crepr{y}^{-1}}{\Xcal})\:g(\coset{\crepr{y}}{H\backslash G/K}).
\vspace{-5pt}
\end{equation} 
}
\bigskip

\ignore{
\begin{defn}\label{def: convo homo3}
Let \m{G} be a finite or countable group, \m{H} and \m{K} be two subgroups of \m{G},\:  
\m{f\colon G/H\<\to\CC}, and \m{g\colon H\backslash G/K\<\to\CC}. 
We then define the \df{convolution} of \m{f} with \m{g} as 
\m{f\ast g\colon G/K\to\CC} with  
\begin{equation}\label{eq: convo homo3}
(f\ast g)(x)=\abs{H} \sum_{y\in H\backslash G} f(\coset{\crepr{x} \crepr{y}^{-1}}{\Xcal})\:g(\coset{\crepr{y}}{H\backslash G/K}).
\end{equation} 
As before, the \m{\abs{H}} scaling factor appears in this equation for compatibility with \rf{eq: convo gp0}, 
and when \m{H} is infinite, it may be dropped.   
\end{defn}
}

\noindent 
Since \m{f\mapsto f\ast g} is a map from one homogeneous space, 
\m{\Xcal=G/H}, to another homogeneous space, \m{\Ycal=H/K}, 
it is this last defintion that will be of most relevance to us in constructing neural networks. 


\ignore{
\subsection{Convolution of vector valued functions}

Since neural nets have multiple channels, we need to further 
extend Equations \ref{eq: convo gp0}, \ref{eq: convo gp1} and Definitions \ref{def: convo homo1}--\ref{def: convo homo3} to vector/matrix valued functions. 
Once again, there are multiple cases to consider. 
\\

\newcommand{\Hom}{\mathop{\textrm{Hom}}}

\begin{defn}\label{def: convo homov}
Let \m{G} be a finite or countable group, and \m{\Xcal} and \m{\Ycal} be (left or right) quotient spaces of \m{G}. 
\setlength{\plitemsep}{3pt}
\begin{compactenum}[~1.] 
\item If \m{f\colon \Xcal\to \CC^m}, and \m{g\colon \Ycal\to \CC^m}, we define \m{f\ast g\colon G\to\CC} with 
\vspace{-7pt}  
\begin{equation}\label{eq: convo homov0}
(f\ast g)(u)=\sum_{v\in G} f\tup^G(uv^{-1})\cdot g\tup^G(v), 
\vspace{-12pt}
\end{equation}
where \m{\cdot} denotes the dot product.  
\item If \m{f\colon \Xcal\to \CC^{n\times m}}, and \m{g\colon \Ycal\to \CC^m}, we define \m{f\ast g\colon G\to\CC^n} with  
\vspace{-7pt}
\begin{equation}\label{eq: convo homo1}
(f\ast g)(u)=\sum_{v\in G} f\tup^G(uv^{-1})\,\times\,  g\tup^G(v), 
\vspace{-7pt}
\end{equation}
where \m{\times} denotes the matrix/vector product.   
\item If \m{f\colon \Xcal\to \CC^{m}}, and \m{g\colon \Ycal\to \CC^{n\times m}}, we define \m{f\ast g\colon G\to\CC^m} with 
\vspace{-7pt}
\begin{equation}\label{eq: convo homo2}
(f\ast g)(u)=\sum_{v\in G}   f\tup^G(uv^{-1})\: \tilde\times\: g\tup^G(v), 
\vspace{-7pt}
\end{equation}
where \m{\V v \tilde\times A} denotes the ``reverse matrix/vector product'' \m{A\V v}.   
\end{compactenum}
Since in cases 2 and 3 the nature of the product is clear from the definition of \m{f} and \m{g}, 
we will omit the \m{\times} and \m{\tilde \times} symbols. 
The specializations of these formulae to the cases of Definitions 
\ref{def: convo homo1}--\ref{def: convo homo3} are as to be expected.  
\end{defn}
}



\ignore{
Therefore, we now discuss the generalization of convolution to the case when either \m{f} or \m{g} or 
both are functions on a homogeneous space \m{\Xcal=G/H}. 

The main device that we need for this generalization is the correspondence between left cosets in \m{G} 
and elements of \m{\Xcal} established in Section \ref{sec: homo}, specifically the  
mapping \m{G\to \Xcal} taking \m{u\mapsto u(x_0)}.  
Recall that we introduced the notation \m{\coset{u}{\Xcal}} for \m{u(x_0)}, and \m{\crepr{x}} 
for the representative of the coset corresponding to \m{x}. 
}
\ignore{

\ignore{
It turns out that convolution also generalizes  to the case when at least one of the two 
functions \m{f} and \m{g} (without loss of generality, we will take \m{g}) is a function not on \m{G} 
itself but on a homogeneous space \m{\Xcal} of \m{G}. 
The key to this further generalization is our observation in Section \ref{sec: homo}  
that fixing some \m{x_0\tin\Xcal} as the origin establishes a natural mapping \m{G\to\Xcal}, 
taking each group element \m{u\tin G} to \m{u(x_0)\tin \Xcal}. 
Because this mapping will be so fundamental to the following, we introduce the shorthand 
\m{[u]_\Xcal\!:=\!u(x_0)} to denote it, hiding the dependence on the (arbitrary) choice of \m{x_0}. 
Note that if \m{v} is some other element of \m{G}, then \m{v([u]_\Xcal)=vu(x_0)=[vu]_\Xcal}, 
which is an identity that we will make use of in the following. 
\\
\framedef{
\begin{defn}\label{def: convo gp}
Let \m{G} be a compact group, \m{\Xcal} be a homogeneous space of \m{G}, \m{f\colon G\to\CC} and 
\m{g\colon \Xcal\to\CC}. 
Then we define the \df{convolution} of \m{f} with \m{g} as 
\begin{equation}\label{eq: convo gp}
(f\ast g)(u)=\int_G f(uv^{-1})\,g([v]_{\Xcal})\:d\mu(v), 
\end{equation}
where \m{\mu} is the Haar measure on \m{G}. In the case that \m{G} is finite or infinite but 
countable, \rf{eq: convo homo1} reduces to 
\begin{equation*}
(f\ast g)(u)=\sum_{v\in G} f(uv^{-1})\,g([v]_\Xcal).
\end{equation*} 
Note that while \m{g} is a function on \m{\Xcal},\: \m{f\ast g} is still a function on the full group, \m{G}.  
\end{defn}
}
\bigskip 
}

It turns out that convolution also generalizes  to the case when at least one of the two 
functions \m{f} and \m{g} (without loss of generality, we will take \m{f}) is a function not on \m{G} 
itself but on a homogeneous space \m{\Xcal} of \m{G}. 
The key to this generalization is the observation in Section \ref{sec: homo}  
that fixing some ``origin'' \m{x_0} for \m{\Xcal} establishes a natural mapping from \m{G\to\Xcal}, 
taking \m{u\mapsto u(x_0)}. 
Recall that we introduced the shorthand \m{\coset{u}{\Xcal}} for \m{u(x_0)}. 

To simplify our exposition, in the following we only cite convolution formulae for 
finite/countable groups. The generalization to compact groups is straightforward, but not strictly 
needed for our disucssion of equivariant neural networks. 
\\

\framedef{
\begin{defn}\label{def: convo homo1}
Let \m{G} be a countable group, \m{\Xcal} a homogeneous space of \m{G}, \m{f\colon \Xcal\to\CC} and 
\m{g\colon G\to\CC}. 
Then we define the \df{convolution} of \m{f} with \m{g} as 
\begin{equation}\label{eq: convo homo1}
f\ast g\colon G\to\CC\qqquad\qqquad  
(f\ast g)(u)=\sum_{v\in G} f([uv^{-1}]_\Xcal)\,g(v).
\vspace{-5pt}
\end{equation} 
\end{defn}
}
\bigskip

\noindent 
Note that \rf{eq: convo homo1} only differs from \rf{eq: convo gp} in the presence of the 
\m{\coset{\,\cdot\,}{\Xcal}} operator. Also note that while \m{f} lives on \m{\Xcal}, both  
\m{g} and \m{f\<\gconv g} are functions on \m{G} itself. 
The following definition further generalizes \rf{eq: convo gp} to the case 
when \m{g} is a function on a second homogeneous space, \m{\Ycal}, which will be the typical 
case in multilayer neural networks.
\\ 

\framedef{
\begin{defn}\label{def: convo homo2}
Let \m{G} be a countable group, \m{\Xcal} and \m{\Ycal} be homogeneous spaces of \m{G},  
\m{f\colon \Xcal\to\CC} and \m{g\colon \Ycal\to\CC}. 
Then we define the \df{convolution} of \m{f} with \m{g} as 
\begin{equation}\label{eq: convo homo2}
f\ast g\colon \Ycal\to\CC\qqquad\qqquad  
(f\ast g)(x)=\sum_{y\in \Ycal} f([\crepr{x}\crepr{y}^{-1}]_\Xcal)\,g(y).
\vspace{-5pt}
\end{equation} 
\end{defn}
}
\bigskip

\noindent 
Note that \rf{eq: convo homo2} features the coset representatives \m{\crepr{x}} and \m{\crepr{y}}.  
At first sight we may worry that this is potentially problematic because the choice of coset representatives 
is arbitrary. Proposition XXX in the Appendix, however, shows that the result of the formula is, 
in fact, invariant to the choice of representatives.

\begin{exmpl} Convolution on the sphere with a Gaussian filter
\end{exmpl}

\subsection{Reduced convolution formulae}\label{sec: reduced}

The shrewd reader might have observed that there is redundancy in formulae \rf{eq: convo homo1} 
and \rf{eq: convo homo2} stemming from the mismatch in the cardinalities of \m{\Xcal} and \m{G}  
(or \m{\Ycal}, in the case of \rf{eq: convo homo2}). 
Specifically, for any \m{v_1,v_2\tin G} related by \m{v_2=h v_1} 
with \m{h\tin H},\: \m{uv_2^{-1}=u v_1^{-1} h^{-1}}, therefore \m{\coset{u v_1^{-1}}{\Xcal}=\coset{u v_2^{-1}}{\Xcal}}, 
and so the corresponding terms in the sum 
reference the same element \m{f(\coset{u v_1^{-1}}{\Xcal})} of \m{f}. 
Recalling from Section \ref{sec: homo} that the set of right \m{H}--cosets \m{Hv=\setof{hv}{h\tin H}} 
is denoted \m{H\backslash G}, \rf{eq: convo homo1} can hence be factored as 
\[
(f\ast g)(u)=\sum_{v\in H\backslash G} f([uv^{-1}]_\Xcal) \sum_{h\in H} g(hv), 
\]
where \m{v\in H\backslash G} denotes that we take one element (the coset representative) from 
each right \m{H}--coset. 
Equivalently, 
\rf{eq: convo homo1} can be rewritten as 
\[
(f\ast g)(u)=\sum_{y\in H\backslash G} f([u\crepr{y}^{-1}]_\Xcal)\, \wbar{g}(y), 
\]
where now \m{y} is a bona fide member of \m{H\backslash G}, and \m{\wbar{g}\colon H\backslash G\to\CC} 
is a function that 
simply sums \m{g} over each right \m{H}--coset:
\[
\wbar{g}(y)=\sum_{h\in H} g(h\crepr{y}). 
\]
Hence, an alternative, reduced form of convolutions for functions on \m{\Xcal}, expressing essentially the 
same as \rf{eq: convo homo1} is the following. 
\\

\framedef{
\begin{defn}\label{def: convo homo1r}
Let \m{G} be a countable group, \m{\Xcal=G/H} a homogeneous space of \m{G} for some subgroup \m{H} of \m{G}, 
\m{f\colon \Xcal\to\CC} and \m{g\colon H\backslash G\to\CC}. 
Then we define the \df{convolution} of \m{f} with \m{g} as 
\begin{equation}\label{eq: convo homo1r}
f\ast g\colon G\to\CC\qqquad\qqquad  
(f\ast g)(u)=\sum_{y\in H\backslash G} f([u\crepr{y}^{-1}]_\Xcal)\:g(y).
\vspace{-5pt}
\end{equation} 
\end{defn}
}
\bigskip
}

\ignore{
The only thing that we need to be careful about is the choice of measure, \m{\mu}. While on \m{\RR^d} 
without further qualification we took this to be the Lebesgue measure, defining the appropriate measure on 
groups can be more complicated. Fortunately, for compact groups there is a canonical choice, called the 
Haar measure, which is the essentially unique (up to rescaling by a constant) 
measure on \m{G} that is invariant to translation in the sense that 
\[\int_G f(u)\,d\mu(u)=\int_G f(t^{-1} u)\:d\mu(u) \]
for any integrable function \m{f} and any \m{t\tin G}. 
Naturally, the class of compact groups encompasses all finite and countable groups. In these cases,  
we take \m{\mu(u)=1} for all \m{u\tin G}, reducing \rf{eq: convo gp0} to 
\begin{equation}\label{eq: convo gp0b}
(f\ast g)(u)=\sum_{v\in G} f(u v^{-1})\,g(v).
\end{equation} 
Note that \rf{eq: conv zd} is a special case of this formula with \m{G=\ZZ^d}, while \rf{eq: conv Znd} is a special 
case with \m{G=\ZZ_n^d}, where \m{\ZZ_n=\cbr{\zton{n-1}}} is the so-called cyclic group of order \m{n}. 
The class of compact groups also includes continuous groups such as the three dimensional rotation group 
\m{\SO(3)} mentioned in Example \ref{ex: sphere} of Section \ref{sec: actions}. The \m{\RR^d} groups are 
not compact, but they are special cases for which there is still a canonical measure, namely the 
Lebesgue measure as mentioned above. An example of a non-compact group where the choice of \m{\mu} 
might get complicated is the Euclidean group \m{\mathop{\textrm{ISO}}(2)} encompassing translations,  
rotations and reflections of the plane. However, we will not need to concern ourselves with such cases in 
this paper. In summary, we have the following definition. 
\\

\begin{defn}\label{def: convo gp}
Let \m{G} be a compact group and \m{f,g\colon G\to \CC}. Then the \df{convolution} of \m{f} with \m{g} 
is defined 
\begin{equation}\label{eq: convo gp}
(f\ast g)(g)=\int_G f(uv^{-1})\,g(v)\:d\mu(v),
\end{equation}
where \m{\mu} is the Haar measure on \m{G}. In the case that \m{G} is finite or infinite but 
countable, \rf{eq: convo gp} reduces to 
\begin{equation}\label{eq: convo gp0}
(f\ast g)(u)=\sum_{v\in G} f(uv^{-1})\,g(v).
\end{equation}
Note that in both cases \m{f\ast g} is a function \m{G\to\CC}.  
\end{defn}
\bigskip

}

\ignore{
As we have seen in Section \ref{sec: actions}, the more common and more 
interesting case in learning problems is when a transformation group acts on functions on homogeneous spaces, 
rather than on functions on the group itself. 
Fortunately, the notion of induced functions, introduced in Section \ref{sec: homo}, allows us to 
immediately generalize convolution to this case. 
\\
}
\ignore{
A closely related concept is that of cross-correlation. Again given \m{f,g\colon \RR\to\RR}, 
the \df{cross-correlation} of \m{f} with \m{g} is \label{p: corr}
\begin{equation*}
(f\star g)(x)=\int f(y)\,g(x\<+y)\,dy,
\end{equation*}
which can be interpretated as template matching: given a template or pattern 
\m{f} that we wish to find in \m{g}, \m{(f\star g)(x)} expresses to what extent that pattern is present 
at location \m{x}. 

Both convolution and cross-correlation have natural extensions to complex valued functions. 
When \m{f,g\colon \RR\to\CC}, the definition of the former is unchanged (except that the mutiplication 
in the integrand in \rf{eq: convolution} is now of course complex multiplication), while the latter changes to 
\begin{equation*}
(f\star g)(x)=\int f(y)^\ast\,g(x\<+y)\,dy,
\end{equation*}
where \m{{}^{\ast}} denotes complex conjugation. The complex valued form of these formulae 
will be important in the following because the generalization of concepts like group representations,  
which we will make extensive use of, is easiest to discuss when working over \m{\CC}. 

From a mathematical point of view, cross-correlation is closely related to convolution.  
In particular, setting \m{f'(x)\<=f(-x)^\ast}, 
the cross-correlation \m{f\star g} can be expressed as the convolution \m{f'\ast g}. 
This is why in the neural networks community the two concepts are often conflated: 
much of what is referred to as convolution in the literature is technically correlation.   
We will not belabor this point too much, and will develop our theory in terms of convolution alone, 
as defined in \rf{eq: convolution}, since mathematically it is a more canonical and somewhat easier to 
generalize concept than cross-correlation. 
\pskip

Convolution generalizes naturally to functions on \m{\RR^d} in the form 
\begin{equation}\label{eq: convo R}
(f\ast g)(\x)=\int g(\x-\y)\,f(\y)\,d\y,
\end{equation}
where now \m{\x,\y\in\RR^d}. It also easy to write down the discretized form of convolution,  
which is of course what is implemented on computers. Given 
\m{f,g\colon \ZZ^d\to\RR} (or \m{f,g\colon \ZZ^d\to\CC}), 
the discrete convolution of \m{f} with \m{g} is defined 
\begin{equation}\label{eq: conv zd}
(f\ast g)(\x)=\sum_{\y\in\ZZ^d}f(\x-\y)\,g(\y)
\qqquad \qqquad \x\in\ZZ^d.
\end{equation}
In practical numerical problems \m{f} and \m{g} do not have infinitely large support, 
so the sum need only be computed in a finite box. 
The convolution \rf{eq: convo1} cited in Section \ref{sec: classic} in the context of classical 
CNNs, for example, is just the two dimensional 
form of \rf{eq: conv zd} in the case when the support of \m{f} is confined to \m{\iinterval{-d}{d}^2}. 
Alternatively, we can constrain \m{f} and \m{g} 
to the \m{n\<\times n\<\ldots\<\times n} hypercube \m{\iinterval{0}{n\<-1}^d} from the start, and wrap the sum 
around its edges, giving 
\begin{equation}\label{eq: conv Znd}
(f\ast g)(\x)=\sum_{\y\in\iinterval{0}{n-1}^d}f(\llbracket \x-\y\rrbracket_n)\,g(\y),
\end{equation}
where \m{\llbracket\x\rrbracket_n:=(x_1\mathop{\textrm{mod}} n,\ldots,x_d\mathop{\textrm{mod}} n)}.  
We shall see that all of these formulae are special cases of convolution on compact groups. 

\subsection{Convolution on groups}
}

\subsection{Relationship to Fourier analysis}\label{sec: fourier}

The nature of convolution on homogeneous spaces is further explicated by considering its form 
in Fourier space (see \cite{Terras}). 
Recall that the \df{Fourier transform} of a function \m{f} on a countable group is defined 
\begin{equation}\label{eq: gp-ft2}
\h f(\rho_i)=\sum_{u\in G} f(u)\ts \rho_i(u), \qqquad i=0,1,2,\ldots,
\end{equation}
where \m{\rho_0,\rho_1,\ldots} are matrix valued functions called \df{irreducible representations} 
or \df{irreps} of \m{G} (see the Appendix for details). 
As expected, the generalization of this to the case when \m{f} is a function on \m{G/H}, \m{H\backslash G} or 
\m{H\backslash G/K} is 
\begin{equation*}
\h f(\rho_i)=\sum_{u\in G} \rho_i(u)\ts f\tup^{G}(u), \qqquad i=1,2,\ldots.
\end{equation*}
Analogous formulae hold for continuous groups, involving integration with respect to the Haar measure. 

At first sight it might be surprising that the Fourier transform of a function on a quotient space 
consists of the same number of matrices of the same sizes as the Fourier transform of a function on \m{G} 
itself,  since \m{G/H}, \m{H\backslash G} or \m{H \backslash G/K} are smaller objects than \m{G}. 
This puzzle is resolved by the following proposition, which tells us that in the latter 
cases, the Fourier matrices have characteristic sparsity patterns. 
 
\begin{prop}\label{prop: ft quotient}
Let \m{\rho} be an irrep of \m{G}, and assume that on restriction to \m{H} it decomposes into 
irreps of \m{H} in the form \m{\rho\vert_H=\mu_1\oplus \mu_2\oplus \ldots \oplus \mu_k}. 
Let \m{\h f} be the Fourier transform of a function \m{f\colon G/H\to\CC}. 
Then \m{[\h f(\rho)]_{\ast,j}=0} 
unless the block at column \m{j} in the decomposition of \m{\rho\vert_H} is the trivial representation.
Similarly, if \m{f\colon H\backslash G\to\CC}, then \m{[\h f(\rho)]_{i,\ast}=0}
unless the block of \m{\rho\vert_H} at row \m{i} is the trivial representation. 
Finally, if \m{f\colon H\backslash G/K\to\CC}, then \m{[\h f(\rho)]_{i,j}=0} unless 
the block of \m{\rho\vert_H} at row \m{i} is the trivial representation of \m{H} \emph{and} 
the block at column \m{j} in the decomposition of \m{\rho\vert_K} is the trivial representation of \m{K}.
\end{prop}

Schematically, this proposition implies that in the three different cases, 
the Fourier matrices have three different forms of sparsity:
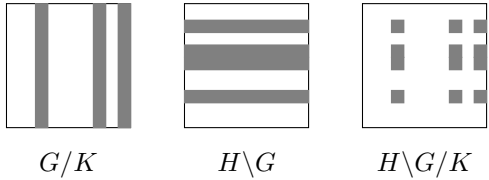
\[
\underset{\displaystyle G/K}{
\underset{\mbox{}}{
\begin{tikzpicture}[scale=0.055]
\draw (0,0) rectangle +(30,-30); 
\filldraw[gray] (7,0) rectangle +(3,-30);
\filldraw[gray] (21,0) rectangle +(3,-30);
\filldraw[gray] (27,0) rectangle +(3,-30);
\end{tikzpicture}
}}
\hspace{20pt}
\underset{\displaystyle H\backslash G}{
\underset{\mbox{}}{
\begin{tikzpicture}[scale=0.055]
\draw (0,0) rectangle +(30,-30); 
\filldraw[gray] (0,-7) rectangle +(30,3);
\filldraw[gray] (0,-13) rectangle +(30,3);
\filldraw[gray] (0,-16) rectangle +(30,3);
\filldraw[gray] (0,-24) rectangle +(30,3);
\end{tikzpicture}
}}
\hspace{20pt}
\underset{\displaystyle H\backslash G/K}{
\underset{\mbox{}}{
\begin{tikzpicture}[scale=0.055]
\draw (0,0) rectangle +(30,-30); 
\filldraw[gray] (7,-7) rectangle +(3,3);
\filldraw[gray] (21,-7) rectangle +(3,3);
\filldraw[gray] (27,-7) rectangle +(3,3);
\filldraw[gray] (7,-13) rectangle +(3,3);
\filldraw[gray] (21,-13) rectangle +(3,3);
\filldraw[gray] (27,-13) rectangle +(3,3);
\filldraw[gray] (7,-16) rectangle +(3,3);
\filldraw[gray] (21,-16) rectangle +(3,3);
\filldraw[gray] (27,-16) rectangle +(3,3);
\filldraw[gray] (7,-24) rectangle +(3,3);
\filldraw[gray] (21,-24) rectangle +(3,3);
\filldraw[gray] (27,-24) rectangle +(3,3);
\end{tikzpicture}
}}
\]

Fortuitously, just like in the classical, Euclidean case, 
convolution also takes on a very nice form in the Fourier domain, even when \m{f} 
or \m{g} (or both) are defined on homogeneous spaces. 

\begin{prop}[Convolution theorem on groups]\label{prop: convo}
Let \m{G} be a compact group, \m{H} and \m{K} subgroups of \m{G}, and 
\m{f,g} be complex valued functions on \m{G}, \m{G/H}, \m{H\backslash G} or \m{H\backslash G/K}. 
In any combination of these cases, 
\begin{equation}\label{eq: convo thm}
\h{f\<\ast g}(\rho_i)=\h f(\rho_i)\, \h g(\rho_i)
\end{equation}
for any given system of irreps \m{\Rcal_G=\cbrN{\seqzi{\rho}}}. 
\end{prop}

Plugging in matrices with the appropriate sparsity patterns into \rf{eq: convo thm} 
now gives us an intuitive way of thinking about Cases I--III above. 

\subsubsection*{\textbf{Case I:~ \m{\Xcal\<=G} and \m{\Ycal\<=G/H}}}

Mutiplying a column sparse matrix with a dense matrix from the left gives a column sparse 
matrix with the same pattern, therefore \m{f\ast g} is a function on \m{G/H}:
\[
\underset{\displaystyle \h{f\ast g}(\rho)}{
\underset{\mbox{}}{
\br{~
\begin{tikzpicture}[baseline=-20, scale=0.04]
\draw (0,0) rectangle +(30,-30); 
\filldraw[gray] (7,0) rectangle +(3,-30);
\filldraw[gray] (21,0) rectangle +(3,-30);
\filldraw[gray] (27,0) rectangle +(3,-30);
\end{tikzpicture}
~}}
}
=
\underset{\displaystyle \h{f}(\rho)}{
\underset{\mbox{}}{
\br{~
\begin{tikzpicture}[baseline=-20, scale=0.04]
\draw (0,0) rectangle +(30,-30); 
\filldraw[gray] (0,0) rectangle +(30,-30);
\end{tikzpicture}
~}}
}
\times 
\underset{\displaystyle \h{g\tupr{G}}(\rho)}{
\underset{\mbox{}}{
\br{~
\begin{tikzpicture}[baseline=-20, scale=0.04]
\draw (0,0) rectangle +(30,-30); 
\filldraw[gray] (7,0) rectangle +(3,-30);
\filldraw[gray] (21,0) rectangle +(3,-30);
\filldraw[gray] (27,0) rectangle +(3,-30);
\end{tikzpicture}
~}}
}
\:.
\]

\subsubsection*{\textbf{Case II:~ \m{\Xcal\<=G/H} ~and~ \m{\Ycal\<=H\backslash G}}}

Multiplying a column sparse matrix from the right by another matrix picks out the corresponding 
rows of the second matrix. Therefore, if \m{f} is a function on \m{G/H}, then w.l.o.g.\:we can take 
\m{g} to be a function on \m{H\backslash G}. 
\[
\underset{\displaystyle \h{f\ast g}(\rho)}{
\underset{\mbox{}}{
\br{~
\begin{tikzpicture}[baseline=-20, scale=0.04]
\draw (0,0) rectangle +(30,-30); 
\filldraw[gray] (0,0) rectangle +(30,-30);
\end{tikzpicture}
~}}
}
=
\underset{\displaystyle \h{f\tupr{G}}(\rho)}{
\underset{\mbox{}}{
\br{~
\begin{tikzpicture}[baseline=-20, scale=0.04]
\draw (0,0) rectangle +(30,-30); 
\filldraw[gray] (7,0) rectangle +(3,-30);
\filldraw[gray] (21,0) rectangle +(3,-30);
\filldraw[gray] (27,0) rectangle +(3,-30);
\end{tikzpicture}
~}}
}
\times 
\underset{\displaystyle \h{g\tupr{G}}(\rho)}{
\underset{\mbox{}}{
\br{~
\begin{tikzpicture}[baseline=-20, scale=0.04]
\draw (0,0) rectangle +(30,-30); 
\filldraw[gray] (0,-7) rectangle +(30,-3);
\filldraw[gray] (0,-21) rectangle +(30,-3);
\filldraw[gray] (0,-27) rectangle +(30,-3);
\end{tikzpicture}
~}}
}
\:.
\]

\subsubsection*{\textbf{Case III:~ \m{f\colon G/H\to\CC} ~and~ \m{g\colon H\backslash G/K\to\CC}}}

Finally, if \m{f} is a function on \m{G/H}, and we want to make \m{f\ast g} to be a function on 
\m{G/K}, then we should take \m{g\colon H\backslash G/K}:
\[
\underset{\displaystyle \h{f\ast g}(\rho)}{
\underset{\mbox{}}{
\br{~
\begin{tikzpicture}[baseline=-20, scale=0.04]
\draw (0,0) rectangle +(30,-30); 
\filldraw[gray] (7,0) rectangle +(3,-30);
\filldraw[gray] (13,0) rectangle +(3,-30);
\filldraw[gray] (16,0) rectangle +(3,-30);
\filldraw[gray] (24,0) rectangle +(3,-30);
\end{tikzpicture}
~}}
}
=
\underset{\displaystyle \h{f\tupr{G}}(\rho)}{
\underset{\mbox{}}{
\br{~
\begin{tikzpicture}[baseline=-20, scale=0.04]
\draw (0,0) rectangle +(30,-30); 
\filldraw[gray] (7,0) rectangle +(3,-30);
\filldraw[gray] (21,0) rectangle +(3,-30);
\filldraw[gray] (27,0) rectangle +(3,-30);
\end{tikzpicture}
~}}
}
\times 
\underset{\displaystyle \h{g\tupr{G}}(\rho)}{
\underset{\mbox{}}{
\br{~
\begin{tikzpicture}[baseline=-20, scale=0.04]
\draw (0,0) rectangle +(30,-30); 
\filldraw[gray] (7,-7) rectangle +(3,3);
\filldraw[gray] (7,-21) rectangle +(3,3);
\filldraw[gray] (7,-27) rectangle +(3,3);
\filldraw[gray] (13,-7) rectangle +(3,3);
\filldraw[gray] (13,-21) rectangle +(3,3);
\filldraw[gray] (13,-27) rectangle +(3,3);
\filldraw[gray] (16,-7) rectangle +(3,3);
\filldraw[gray] (16,-21) rectangle +(3,3);
\filldraw[gray] (16,-27) rectangle +(3,3);
\filldraw[gray] (24,-7) rectangle +(3,3);
\filldraw[gray] (24,-21) rectangle +(3,3);
\filldraw[gray] (24,-27) rectangle +(3,3);
\end{tikzpicture}
~}}
}
\:.
\]

\ignore{
There are two main reasons why this theory is important for our exposition: 
(a) because generalized Fourier transforms help illuminate the definitions of convolution on quotient spaces 
presented in the previous section; 
(b) because they a key tool for 
proving our main theorem in Section \ref{sec: gcn1}. 
However, 
a reader who is only interested in the main ideas of this paper at a high level  
may skip the rest of this section in a first pass. 
\pskip
%
The base case of generalized Fourier transforms requires only 
two basic facts from representation theory: (a) that a representation 
of a compact group \m{G} is a complex matrix valued function \m{\rho\colon G\to\CC^{d_\rho\times d_\rho}} that 
obeys \m{\rho(uv)=\rho(u)\ts\rho(v)} for any \m{u,v\tin G}; 
(b) that \m{G} has an (essentially unique) sequence of special representations 
\m{\Rcal_G=\cbrN{\seqi{\rho}}} called \emph{irreps} that 
all other representations are constructed from (see Appendix B). 
The \df{Fourier transform} of \m{f\colon G\to\CC} \ig{(w.r.t. \cbrN{\seqi{\rho}}) }is then defined  
}
\ignore{
Despite \m{\RR} not being a compact group, \rf{eq: Fourier classical} can be seen as a 
special case of \rf{eq: gp-ft}, since \m{e^{-2\pi \iota kx}} trivially obeys 
\m{e^{-2\pi \iota k(x_1+x_2)}=e^{-2\pi \iota kx_1} e^{-2\pi \iota kx_2}}, and 
the functions \m{\rho_k(x)=e^{-2\pi \iota kx}} are, in fact, 
the irreducible representations of \m{\RR}. 
The fundamental novelty in \rf{eq: gp-ft} and \rf{eq: gp-ft2} 
compared to \rf{eq: Fourier classical}, however, is that since, in general 
(in particular, when \m{G} is not commutative),   
irreducible representations are matrix valued functions, 
each ``Fourier component'' \m{\h f(\rho)} is now a matrix. 
In other respects, Fourier transforms on groups behave very similarly to classical Fourier transforms. 
For example, we have an inverse Fourier transform 
\[f(u)=\ovr{\abs{G}}\sum_{\rho\in\Rcal} d_\rho\,\tr\sqbbig{f(\rho)\ts \rho(u)^{-1}},  
\]
and also an analog of the convolution theorem, \rf{eq: convo thm}. 
\\
}
\ignore{
\noindent
\begin{pf}
Let us assume that \m{G} is countable. Then 
\begin{multline*}
\h{f\<\ast g}(\rho_i)=
\sum_{u\in G} \sqbBig{\sum_{v\in G} f(u v^{-1})\,g(v)}\rho_i(u)=
\sum_{u\in G} \sum_{v\in G} f(u v^{-1})\,g(v) \rho_i(uv^{-1})\rho_i(v)=\\
\sum_{v\in G} \sum_{u\in G} f(u v^{-1})\,g(v) \rho_i(uv^{-1})\rho_i(v)=
\sum_{v\in G} \sqbBig{\sum_{u\in G} f(u v^{-1})\,\rho_i(uv^{-1})}g(v) \rho_i(v) =\\
\sum_{v\in G} \sqbBig{\sum_{w\in G} f(w)\,\rho_i(w)}g(v) \rho_i(v) =
\sqbBig{\sum_{w\in G} f(w)\,\rho_i(w)} \sqbBig{\sum_{v\in G} g(v) \rho_i(v)} =
\h f(\rho_i)\, \h g(\rho_i).
\end{multline*}
The continuous case is proved similarly but with integrals with respect Haar measure 
instead of sums. 
\end{pf}
\bigskip 
}
\ignore{
\noindent 
The correlation theorem has a similar analog 
\m{\h {f\star g}(\rho)=\h f(\rho)\,\h g(\rho)^\dag} if we define 
cross-correlation on groups as \m{(f\star g)(u)=\int f(uv)\, g(v)^\ast\:d\mu(v)}. 

\subsection{Fourier transforms on quotient spaces}
}
\ignore{
\pskip
First, recall that if \m{H} is a subgroup of \m{G}, a function \m{f\colon G\to\CC} is called 
\df{right} \m{H}--\df{invariant} if \m{f(uh)=f(u)} for all \m{h\tin H} and all \m{u\tin G}, and it is called 
\df{left} \m{H}--\df{invariant} if \m{f(hu)=f(u)} for all \m{h\tin H} and all \m{u\tin G}. 
We start with two simple lemmas.

\begin{lem}\label{lem: homo-inv}
Let \m{H} and \m{K} be two subgroups of a group \m{G}. Then
\begin{compactenum}[~~1.]
\item If \m{f\colon G/H\to\CC}, then \m{f\tup^G\colon G\to\CC} is right \m{H}--invariant. 
\item If \m{f\colon H\backslash G\to\CC}, then \m{f\tup^G\colon G\to\CC} is left \m{H}--invariant. 
\item If \m{f\colon K\backslash G/H\to\CC}, then \m{f\tup^G\colon G\to\CC} is right \m{H} invariant and 
left \m{K}--invariant. 
\end{compactenum}
\end{lem}

\begin{lem}\label{lem: ft restr}
Let \m{\rho} be an irreducible representation of a countable group \m{G}. 
Then \m{\sum_{u\in G}\rho(u)=0} unless \m{\rho} is the trivial  
representation, \m{\rho_{\tr}(u)=(1)}. 
\end{lem}

\noindent
\begin{pf}
Let us define the functions 
\m{r^\rho_{i,j}(u)=[\rho(u)]_{i,j}}. 
Recall that for \m{f,g\colon G\to\CC}, the inner product \m{\inp{f,g}} is defined 
\m{\inp{f,g}=\sum_{u\in G}f(u)^\ast g(u)}. 
The Fourier transform of a function \m{f} can then be written element-wise as 
\m{[\h f(\rho)]_{i,j}=\inpN{r^\rho_{i,j}^\ast,f}}.
However, since the Fourier transform is a unitary transformation, 
for any \m{\rho,\rho'\tin\Rcal_G}, unless \m{\rho\<=\rho'}, \m{i\<=i'} and \m{j\<=j'}, 
we must have \m{\inpN{r^{\rho}_{i,j},r^{\rho'}_{i',j'}}=0}. 
In particular, 
\begin{mequation*}
\sqbbig{\sum_{u\in G}\rho(u)}_{i,j}=\inpN{r^{\rho_{\tr}}_{1,1},r^{\rho}_{i,j}}=0,
\end{mequation*}
unless \m{\rho=\rho_{\tr}} (and \m{i\<=j\<=1}). 
\end{pf}

\noindent 
Now recall that given an irrep \m{\rho} of \m{G}, the \emph{restriction} of \m{\rho} to \m{H} is 
\m{\rho\vert_H\colon H\to\CC^{d_\rho\times d_\rho}}, where \m{\rho\vert_H(h)=\rho(h)} for all \m{h\tin H}. 
It is trivial to check that \m{\rho\vert_H} is a representation of \m{H}, but, in general, it is not 
irreducible. Thus, by the Theorem of Complete Decomposability (Appendix B), 
it must decompose in the form \m{\rho\vert_H(h)=Q(\mu_1(h)\oplus \mu_2(h)\oplus \ldots \oplus \mu_k(h))Q^\dag} 
for some sequence \sseq{\mu}{k} of irreps of \m{H} and some unitary martrix \m{Q}. 
In the special case when the irreps of 
\m{G} and \m{H} are adapted to \m{H\leq G}, however, \m{Q} is just the unity. 
This is the case that we consider in the following lemmas.  

\begin{lem}\label{lem: ft homo1}
Let \m{H} be a subgroup of a countable group \m{G} and \m{\rho} be an irreducible representation 
of \m{G}. Assume that the restriction of \m{\rho} to \m{H} 
decomposes into a direct sum of irreducible representations of 
\m{H} in the form \m{\rho\vert_H=\mu_1\oplus \mu_2\oplus \ldots \oplus \mu_k}. Let \m{f\colon G/H\to\CC}, and 
the Fourier transform of \m{f} be defined as in Definition \ref{def: FT homo}. Then 
\begin{mequation}
[\h f(\rho)]_{\ast,j}=0
\end{mequation}
unless the block at column \m{j} in the decomposition of \m{\rho\vert_H} is the trivial representation. 
\end{lem}
\noindent 
\begin{pf}
The fact that any \m{u\tin G} can be written uniquely as \m{u\<=gh} where \m{g} is the representative of 
one of the \m{gH} cosets and \m{h\tin H} immediately tells us that \m{\h f(\rho)} factors as
\vspace{-5pt} 
\begin{multline}\label{eq: ft-homo1-fact}
\h f(\rho)=\sum_{u\in G} f\tup^{G}(u)\ts \rho(u)=
\sum_{x\in G/H} \sum_{h\in H} f\tup^{G}(\crepr{x}h)\ts \rho(\crepr{x}h)=
\sum_{x\in G/H} \sum_{h\in H} f(x)\ts \rho(\crepr{x}h)=\\
\sum_{x\in G/H} \sum_{h\in H} f(x)\ts \rho(\crepr{x})\rho(h)=
\sum_{x\in G/H} f(x)\ts \rho(\crepr{x}) \sqbBig{\sum_{h\in H}\rho(h)}.
\end{multline}
However, \m{\rho(h)=\mu_1(h)\oplus \mu_2(h)\oplus \ldots \oplus \mu_k(h)} for some sequence 
of irreps \m{\sseq{\mu}{k}} of \m{H}, so 
\[\sum_{h\in H}\rho(h)=
\sqbBig{\sum_{h\in H}\mu_1(h)}\oplus 
\sqbBig{\sum_{h\in H}\mu_2(h)}\oplus \ldots \oplus  
\sqbBig{\sum_{h\in H}\mu_k(h)}, 
\]
and by Lemma \ref{lem: ft restr} each of the terms in this sum where \m{\mu_i} is \emph{not} the 
trivial representation (on \m{H}) is a zero matrix, zeroing out all the corresponding 
columns in \m{\h f(\rho)}.  
\end{pf}

\noindent 
\input{fig-FThomo}


\begin{lem}\label{lem: ft homo2}
Let everything be as in Lemma \ref{lem: ft homo1}, except that now \m{f\colon H\backslash G\to\CC}. 
Then 
\begin{mequation}
[\h f(\rho)]_{i,\ast}=0
\end{mequation}
unless the block of \m{\rho\vert_H} at row \m{i} is the trivial representation. 
\end{lem}
\noindent 
\begin{pf}
Analogous to the proof of Lemma \rf{lem: ft homo1}, using \m{u=hg} and a 
factorization similar to \rf{eq: ft-homo1-fact} except that \m{\sum_{h\in H}\rho(h)} 
will now multiply \m{\sum_{x\in H\backslash G} f(x)\ts \rho(\crepr{x})} from the left.  
\end{pf}

\begin{lem}\label{lem: ft homo3}
Let everything be as in Lemma \ref{lem: ft homo1}, but now let \m{K} be a second subgroup of \m{G}, 
and assume that \m{\rho\vert_K} decomposes into irreducibles of \m{K} in the form 
\m{\rho\vert_K=\nu_1\oplus \nu_2\oplus \ldots \oplus \nu_p}. 
Let \m{f\colon H\backslash G/K\to\CC}. Then 
\begin{mequation}
[\h f(\rho)]_{i,j}=0
\end{mequation} 
unless both the block of \m{\rho\vert_K} at column \m{j} is the trivial representation \emph{and} 
the block of \m{\rho\vert_H} at row \m{i} is the trivial representation. 
\end{lem}
\noindent 
\begin{pf}
Immediate from combining case 3 of Lemma \ref{lem: homo-inv} with Lemmas \ref{lem: ft homo1} and \ref{lem: ft homo2}. 
\end{pf}

\noindent 
Lemmas \ref{lem: ft homo1}--\ref{lem: ft homo3} hold when the irreps of \m{G} are 
adapted to the irreps of \m{H}. In the more general case, when the \m{Q} matrix in 
\m{\rho\vert_H(h)=Q(\mu_1(h)\oplus \mu_2(h)\oplus \ldots \oplus \mu_k(h))Q^\dag} is 
nontrivial, it is not \m{\h f(\rho)}, but \m{Q^\dag \h f(\rho) Q} that will 
exhibit the row or column sparse structure described in our lemmas.  
However, \m{\h f(\rho)} still inherits the corresponding rank constraints. 

\begin{lem}
Let \m{G} be a compact group, \m{H} and \m{K} two subgroups of \m{G}, and \m{\rho} an irreducible 
representation of \m{G}. 
Let \m{m_{\rho\vert_H}\nts(\mu_\tr)} denote the multipliticity of the trivial representation in the decomposition 
of \m{\rho\vert_H} into a sum of irreps of \m{H}, and \m{m_{\rho\vert_K}\nts(\nu_\tr)} 
be similarly defined for \m{\rho\vert_K}. 
Then the following hold:
\begin{compactenum}
\item If \m{f\colon G/H\to \CC}, then \m{\Trank(\h f(\rho))\<\leq m_{\rho\vert_H}\nts(\mu_\tr)}.
\item If \m{f\colon H\backslash G\to \CC}, then \m{\Trank(\h f(\rho))\<\leq m_{\rho\vert_H}\nts(\mu_\tr)}.
\item If \m{f\colon H\backslash G/K\to \CC}, then 
\m{\Trank(\h f(\rho))\<\leq \min\cbrN{m_{\rho\vert_H}\nts(\mu_\tr),m_{\rho\vert_K}\mts(\nu_\tr)}}.
\end{compactenum}
\end{lem}
\smallskip

\noindent
As promised, we now examine the consequences of Lemmas \ref{lem: ft homo1}--\ref{lem: ft homo3} 
for the three cases of convolution on homogeoneous spaces discussed in Section \ref{sec: conv homo}. 
}

\section{Main result: the connection between convolution and equivariance}\label{sec: gcn1}

We are finally in a position to define the notion of generalized convolutional networks,  
and state our main result connecting convolutions and equivariance. 
\bigskip 

\framedef{
\begin{defn}
Let \m{G} be a compact group and \m{\Ncal} an \m{L\<+1} layer feed-forward network 
in which the \m{i}'th index set is \m{G/H_i} for some subgroup \m{H_i} of \m{G}. 
We say that \m{\Ncal} is a \m{G}\df{--convolutional neural network} (or \df{\m{\mathbf G}-CNN} for short) 
if each of the linear maps \m{\sseq{\phi}{L}} in \m{\Ncal} is a generalized convolution 
(see Definition \ref{def: convo homo0}) of the form  
\[\phi_\ell(f_{\ell-1})=f_{\ell-1}\ast \chi_\ell\]
with some filter \m{\chi_\ell\tin\lins{H_{\ell-1}\backslash G/H_{\ell}}{V_{\ell-1}\times V_\ell}}. 
\end{defn}
}
\smallskip

\framedef{
\begin{thm}\label{thm: GCNN}
Let \m{G} be a compact group and \m{\Ncal} be an \m{L+1} layer feed-forward neural network in which the 
\m{\ell}'th index set is of the form \m{\Xcal_\ell=G/H_\ell}, where \m{H_\ell} is some subgroup of \m{G}. 
Then \m{\Ncal} is equivariant to the action of \m{G} in the sense of Definition \ref{def: equivariant NN} 
if and only if it is a \m{G}-CNN. 
\end{thm}
}
\bigskip

\noindent
Proving this theorem in the forward direction is relatively easy and only requires some elementary 
facts about cosets and group actions. 

\noindent 
\begin{pfof}{Theorem \ref{thm: GCNN} (forward direction)} 
Assume that we translate \m{f_{\ell-1}} by some group element \m{g\tin G} and get \m{f'_{\ell-1}}, i.e., 
\m{f_{\ell-1}'=\Tact^{\ell-1}_g(f_{\ell-1})}, where   
\m{f_{\ell-1}'(x)=f_{\ell-1}(g^{-1}x)}. 
Then \vspace{-6pt}   
\begin{align*}
\phi_{\ell}(f'_{\ell-1})(u)
& =(f'_{\ell-1}\ast \chi_{\ell})(u) \\
& =\sum_{v\in G} f'_{\ell-1}([uv^{-1}]_\Xcal)\,\chi_{\ell}(v) \\
& =\sum_{v\in G} f_{\ell-1}(g^{-1}([uv^{-1}]_\Xcal))\,\chi_{\ell}(v). \vspace{-4pt}
\end{align*}
By \m{g^{-1}([uv^{-1}]_\Xcal)=[g^{-1}uv^{-1}]_{\Xcal}} this is further equal to \vspace{-4pt}
\begin{multline*}
\sum_{v\in G} f_{\ell-1}([g^{-1}uv^{-1}]_\Xcal)\,\chi_{\ell}(v) \\
=(f_{\ell-1}\ast \chi_{\ell})(g^{-1}u) 
=\phi_{\ell}(f_{\ell-1})(g^{-1} u).\vspace{-4pt}
\end{multline*}
Therefore, \m{\phi_{\ell}(f_{\ell-1})} is equivariant with \m{f_{\ell-1}}. 
Since \m{\actfn_\ell} is a pointwise operator, so is \m{f_\ell=\actfn_\ell(\phi_\ell(f_{\ell-1}))}.  
By induction on \m{\ell}, using the transitivity of equivariance,  
this implies that every layer of \m{\Ncal} is equivariant with layer \m{0}. 
Note that this proof holds not only in the base case, when each \m{f_\ell} is a 
function \m{\Xcal\to\CC}, but also in the more general case when \m{f_\ell\colon\Xcal_\ell\to V_\ell} 
and the filters are \m{\chi_\ell\colon \Xcal_\ell\to V_{\ell-1}\times V_\ell}.  
\end{pfof}

\noindent 
Proving the ``only if'' part of Theorem \ref{thm: GCNN} is more technical, therefore we 
leave it to the Appendix.

\ignore{  
requires concepts from representation theory (Appendix B) and the notion of generalized Fourier transforms introduced in Section \ref{sec: fourier}. 
We also need two versions of Schur's Lemma. 

\begin{lem}\textbf{(Schur's lemma I)}
Let \m{\cbrN{\rho(g)\colon U\<\to U}_{g\in G}} and \m{\cbrN{\rho'(g)\colon V\<\to V}_{g\in G}} 
be two irreducible representations of a compact group \m{G}. 
Let \m{\phi\colon U\to V} be a linear (not necessarily invertible) mapping that is equivariant with these 
representations in the sense that \m{\phi(\rho(g)(u))=\rho'(g)(\phi(u))} for any \m{u\tin U}. 
Then, unless \m{\phi} is the zero map, \m{\rho} and \m{\rho'} are equivalent representations. 
\end{lem}

\begin{lem}\textbf{(Schur's lemma II)} 
Let \m{\cbrN{\rho(g)\colon U\to U}_{g\in G}} be an irreducible representation of a compact group \m{G} 
on a space \m{U}, and \m{\phi\colon U\to U} a linear map that commutes with each \m{\rho(g)} 
(i.e., \m{\rho(g)\circ \phi=\phi\circ \rho(g)} for any \m{g\tin G}). 
Then \m{\phi} is a multiple of the identity.
\end{lem}

\noindent 
We build up the proof through a sequence of lemmas. 

\begin{lem}\label{lem: GCNN1}
Let \m{U} and \m{V} be two vector spaces on which a compact group \m{G} acts 
by the linear actions \m{\cbrN{T_g\colon U\!\<\to U}_{g\in G}} and \m{\cbrN{T'_g\colon V\!\<\to V}_{g\in G}}, 
respectively. Let \m{\phi\colon U\!\<\to V} be a linear map that is equivariant with the \m{\cbrN{T_g}} 
and \m{\cbrN{T'_g}} actions, and \m{W} be an irreducible subspace of \m{U} (with respect to \m{\cbrN{T_g}}). 
Then \m{Z\<=\phi(W)} is an irreducible subspace of \m{V}, and the restriction of \m{\cbrN{T_g}} to 
\m{W}, as a representation, is equivalent with the restriction of \m{\cbrN{T'_g}} to \m{Z}. 
\end{lem}

\noindent
\begin{pf}
Assume for contradiction that \m{Z} is reducible, i.e., that it has a proper subspace \m{\Zcal\subset Z} 
that is fixed by \m{\cbrN{T'_g}} (in other words, \m{T'_g(v)\tin \Zcal} for all \m{v\tin \Zcal} and \m{g\tin G}). 
Let \m{v} be any nonzero vector in \m{\Zcal}, \m{u\tin U} be such that \m{\phi(u)\<=v},  
and \m{\mathcal{W}=\Tspan\setof{T_g(u)}{g\tin G}}. 
Since \m{W} is irreducible, \m{\mathcal{W}} cannot be a proper subspace of \m{W}, so \m{\mathcal{W}=W}. 
Thus, 
\[Z=\phi(\Tspan\setof{T_g(u)}{g\tin G})=\Tspan\setofN{T'_g(\phi(u))}{g\tin G}=\Tspan\setofN{T'_g(v)}{g\tin G}\subseteq\Zcal,\] 
contradicting our assumption. 
Thus, the restriction \m{\cbrN{T_g\vert_W}} of \m{\cbrN{T_g}} to \m{W} and the restriction 
\m{\cbrN{T'_g\vert_Z}} of \m{\cbrN{T'_g}} to \m{Z} are both irreducible representations, 
and \m{\phi\colon W\to Z} is a linear map that is equivariant with them.  
By Schur's lemma it follows that \m{\cbrN{T_g\vert_W}}  and \m{\cbrN{T'_g\vert_Z}} are 
equivalent representations. 
\end{pf}

\begin{lem}\label{lem: GCNN2}
Let \m{U} and \m{V} be two vector spaces on which a compact group \m{G} acts 
by the linear actions \m{\cbrN{T_g\colon U\<\to U}_{g\in G}} and \m{\cbrN{T'_g\colon V\<\to V}_{g\in G}},    
and let \m{U=\soperi{U}{\oplus}} and \m{V=\soperi{V}{\oplus}}  
be the corresponding isotypic decompositions. 
Let \m{\phi\colon U\to V} be a linear map that is equivariant with the \m{\cbrN{T_g}} and \m{\cbrN{T'_g}} 
actions. Then \m{\phi(U_i)\subseteq V_i} for any \m{i}. 
\end{lem}

\noindent 
\begin{pf}
Let \m{U_i=U_i^1\oplus U_i^2\oplus\ldots } be the decomposition of \m{U_i} into irreducible 
\m{G}--modules, and \m{V_i^j=\phi(U_i^j)}. 
By Lemma \ref{lem: GCNN1}, each \m{V_i^j} is an irreducible \m{G}--module that is equivalent with \m{U_i^j}, 
hence \m{V_i^j\subseteq V_i}. Consequently, \m{\phi(U_i)=\phi(U_i^1\oplus U_i^2\oplus\ldots)\subseteq V_i}.  
\end{pf}
\bigskip

\begin{lem}\label{lem: GCNN3}
Let \m{\Xcal=G/H} and \m{\Xcal'=G/K} be two homogeneous spaces of a compact group \m{G}, 
let \m{\cbrN{\Tact_g\colon L(\Xcal)\<\to L(\Xcal)}_{g\in G}} and 
\m{\cbrN{\Tact'_g\colon L(\Xcal')\<\to L(\Xcal')}_{g\in G}} be 
the corresponding translation actions, and let \m{\phi\colon L(\Xcal)\to L(\Xcal')} be a linear map that 
is equivariant with these actions.  
Given \m{f\tin L(\Xcal)} let \m{\h f} denote its Fourier transform with respect to 
a specific choice of origin \m{x_0\tin \Xcal} and system or irreps \m{\Rcal_G=\cbrN{\seqi{\rho}}}. 
Similarly, \m{\h f'} is the Fourier transform of \m{f'\tin L(\Xcal')}, 
with respect to some \m{x_0'\tin\Xcal'} and the same system of irreps. 

Now if \m{f'=\phi(f)}, then each Fourier component of \m{f'} is a linear function of the corresponding 
Fourier component of \m{f}, i.e., there is a sequence of linear maps \m{\cbrN{\Phi_i}} such that 
\begin{mequation*}
\h{f'}(\rho_i)=\Phi_i(\h f(\rho_i)).
\end{mequation*}
\end{lem}
\noindent
\begin{pf} 
Let \m{\soperi{U}{\oplus}} and \m{\soperi{V}{\oplus}} be the isotypic decompositions of \m{L(\Xcal)} 
and \m{L(\Xcal')} with respect to the \m{\cbrN{\Tact_g}} and \m{\cbrN{\Tact'_g}} actions. 
By our discussion in Section \ref{sec: isotypics}, each Fourier component \m{\h f(\rho_i)} 
captures the part of \m{f} falling in the corresponding isotypic subspace \m{U_i}. 
Similarly, \m{\h f'(\rho_j)} captures the part of \m{f'} falling in \m{V_j}.    
Lemma \ref{lem: GCNN2} tells us that because \m{\phi} is equivariant with the translation actions, 
it maps each \m{U_i} to the corresponding isotypic \m{V_i}. 
Therefore, \m{\h f'(\rho_i)=\Phi_i(\h f(\rho_i))} for some function \m{\Phi_i}. 
By the linearity of \m{\phi}, each \m{\Phi_i} must be linear. 
\end{pf}
\smallskip

\noindent 
Lemma \ref{lem: GCNN3} is a big step towards describing what form equivariant 
mappings take in Fourier space,
but it doesn't yet fully pin down the individual \m{\Phi_i} maps. 
We now focus on a single pair of isotypics \m{(U_i,V_i)} and the corresponding 
map \m{\Phi_i} \ig{\colon \CC^{d\times d}\to \CC^{d\times d}}taking \m{\h f(\rho_i)\mapsto \h f'(\rho_i)}. 
We will say that \m{\Phi_i} is an \emph{allowable} map if there is some equivariant \m{\phi} 
such that \m{\h{\phi(f)}(\rho_i)=\Phi_i(\h f(\rho_i))}.  
Clearly, \ig{by the linearity of equivariance, }if \m{\seqi{\Phi}} are individually allowable, then 
they are also jointly allowable. 
We will use the shorthand \m{\delta=\d_{\rho_i}}. 
\\

\begin{lem}\label{lem: GCNN5}
All linear maps of the form \m{\Phi_i\colon M\mapsto MB} where \m{B\tin\CC^{\delta\times \delta}} are allowable. 
\end{lem}
\noindent 
\begin{pf}
Recall that the \m{\cbrN{\Tact_g}} action takes \m{f\<\mapsto f^g}, where \m{f^g(x)\<=f(g^{-1}x)}. 
In Fourier space, 
\vspace{-5pt}
\begin{multline}\label{eq: GCNNeq1}
\h{f^g}(\rho_i)=
\sum_{u\in G}\rho_i(u)\ts f^g\tup^G\nts(u)=
\sum_{u\in G}\rho_i(u)\ts f\tup^G\nts(g^{-1} u)=
\sum_{w\in G}\rho_i(gw)\ts f\tup^G\nts(w)=\\
\rho_i(g) \sum_{w\in G}\rho_i(w)\ts f\tup^G\nts(w)=
\rho_i(g)\ts \h f(\rho_i).
\vspace{-5pt}
\end{multline}
(This is actually a general result called the (left) translation theorem.) 
Thus, 
\[
\Phi_i\brbig{\h{\Tact_g(f)}(\rho_i)}=
\Phi_i\brbig{\rho_i(g)\h f(\rho_i)}=
\rho_i(g)\ts \h f(\rho_i)\ts B.
\]
Similarly, the \m{\cbrN{\Tact'_g}} action maps 
\m{\h f'(\rho_i)\mapsto g(\rho_i) \h f'(\rho_i)}, so  
\[
\Tact'_g\brbig{\Phi_i(\h f(\rho_i))}=
\Tact'_g\brbig{\h f(\rho_i)\ts B}=
\rho_i(g)\ts \h f(\rho_i)\ts B.
\]
Therefore, \m{\Phi_i} is equivariant with the \m{\cbrN{\Tact}} 
and \m{\cbrN{\Tact'}} actions. 
\end{pf}

\begin{lem}\label{lem: GCNN6}
Let \m{\Phi_i\colon M\mapsto BM} for some \m{B\tin\CC^{\delta\times \delta}}. 
Then \m{\Phi_i} is not allowable unless \m{B} is a multiple of the identity.  
Moreover, this theorem also hold in the columnwise sense that if \m{\Phi_i\colon M\to M'} such 
that \m{[M']_{\ast,j}=B_j\, [M]_{\ast,j}} for some sequence of matrices \m{\sseq{B}{d}}, 
then \m{\Phi_i} is not allowable unless each \m{B_j} is a multiple of the identity. 
\end{lem}
\noindent 
\begin{pf}
Following the same steps as in the proof of Lemma \ref{lem: GCNN5}, we now have 
\begin{gather*}
\Phi_i\brbig{\h{\Tact_g(f)}(\rho_i)}=B\ts \rho_i(g)\ts \h f(\rho_i),\\ 
\Tact'_g\brbig{\Phi_i(\h f(\rho_i))}=\rho_i(g)\ts B\ts \h f(\rho_i).
\end{gather*} 
However, by the second form of Schur's Lemma, we cannot have \m{B\ts \rho_i(g)=\rho_i(g) B} 
for all \m{g\tin G}, unless \m{B} is a multiple of the identity. 
\end{pf}

\newcommand{\alpc}[4]{\alpha_{#1,#2,#3,#4}}
\newcommand{\alpcd}[4]{\alpha'{}_{\hspace{-3pt}#1,#2,#3,#4}}
\newcommand{\alpcdd}[4]{\alpha''{}_{\hspace{-5pt}#1,#2,#3,#4}}
\newcommand{\alpcz}[4]{\alpha^{0}{}_{\hspace{-5pt}#1,#2,#3,#4}}
\begin{lem}\label{lem: GCNN7}
\m{\Phi_i} is allowable if and only if it is of the form \m{M\mapsto MB} for some 
\m{B\tin\CC^{\delta\times \delta}}.
\end{lem}
\noindent 
\begin{pf}
For the ``if'' part of this lemma,  see Lemma \ref{lem: GCNN5}.  
For the ``only if'' part, note that the set of allowable \m{\Phi_i} form a subspace of all linear maps 
\m{\CC^{\delta\times \delta}\to\CC^{\delta\times \delta}}, and any allowable \m{\Phi_i} can be expressed in the form
\[[\Phi_i(M)]_{a,b}=\sum_{c,d} \alpc{a}{b}{c}{d} M_{c,d}.\]  
By Lemma \ref{lem: GCNN6}, if \m{a\neq c} but \m{b\<=d}, then \m{\alpc{a}{b}{c}{d}\<=0}. 
On the other hand, by Lemma \ref{lem: GCNN5} if \m{a\<=c}, then \m{\alpc{a}{b}{c}{d}} 
can take on any value, regardless of the values of \m{b} and \m{d}, as long as 
\m{\alpc{a}{b}{a}{d}} is constant across varying \m{a}. 

Now consider the remaining case \m{a\neq c} and \m{b\neq d}, and assume that   
\m{\alpc{a}{b}{c}{d}\neq 0} while \m{\Phi_i} is still allowable. 
Then, by Lemma \ref{lem: GCNN5}, it is possible to construct a second allowable map \m{\Phi_i'} 
(namely one in which \m{\alpcd{a}{d}{a}{b}=1} and 
\m{\alpcd{a}{d}{x}{y}=0} for all \m{(x,y)\neq(c,d)}) such that in the composite map 
\m{\Phi_i''=\Phi_i'\circ\Phi_i}, \m{\alpcdd{a}{d}{c}{d}\neq 0}. 
Thus, \m{\Phi_i''} is not allowable. However, the composition of one allowable map 
with another allowable map is allowable, contradicting our assumption that \m{\Phi_i} is allowable. 

Thus, we have established that if \m{\Phi_i} is allowable, then \m{\alpc{a}{b}{c}{d}\<=0}, unless \m{a\<=c}. 
To show that any allowable \m{\Phi_i} of the form \m{M\mapsto MB}, 
it remains to prove that additionally \m{\alpc{a}{b}{a}{d}} is constant across \m{a}. 
Assume for contradiction that \m{\Phi_i} is allowable, but for some \m{(a,e,b,d)} indices 
\m{\alpc{a}{b}{a}{d}\neq\alpc{e}{b}{e}{d}}. 
Now let \m{\Phi_0} be the allowable map that zeros out every column except column \m{d} 
(i.e., \m{\alpcz{x}{d}{x}{d}=1} for all \m{x}, but all other coefficients are zero), 
and let \m{\Phi'} be the allowable map that moves column \m{b} to column \m{d} 
(i.e., \m{\alpcd{x}{d}{x}{b}=1} for any \m{x}, but all other coeffcients are zero). 
Since the composition of allowable maps is allowable, we expect \m{\Phi''=\Phi'\circ\Phi\circ \Phi^0} 
to be allowable. 
However \m{\Phi''} is a map that falls under the purview of Lemma \ref{lem: GCNN6}, yet 
\m{\alpcdd{a}{d}{a}{d}\neq\alpcdd{e}{d}{e}{d}} (i.e., \m{M_j} is not a multiple of the identity) 
creating a contradiction. 
\end{pf}

\bigskip 
\noindent 
\begin{pfof}{Theorem \ref{thm: GCNN} (reverse direction)}
For simplicty we first prove the theorem assuming \m{\Ycal_\ell\<=\CC} for each \m{\ell}. 

Since \m{\Ncal} is a G-CNN, each of the mappings  
\m{(\actfn_\ell\circ \phi_\ell)\colon \lins{\Xcal_{\ell-1}}{}\to \lins{\Xcal_{\ell}}{}} 
is equivariant with the corresponding 
translation actions \m{\cbrN{\Tact^{\ell-1}_g}_{g\in G}} and \m{\cbrN{\Tact_g^\ell}_{g\in G}}. 
Since \m{\actfn_\ell} is a pointwise operator, this is equivalent to asserting that 
\m{\phi_\ell} is equivariant with \m{\cbrN{\Tact^{\ell-1}_g}_{g\in G}} and \m{\cbrN{\Tact_g^\ell}_{g\in G}}. 

Letting \m{\Xcal=\Xcal_{\ell-1}} and \m{\Xcal'=\Xcal_\ell}, Lemma \ref{lem: GCNN5} then tells us 
the the Fourier transforms of \m{f_{\ell-1}} and \m{\phi_\ell(f_{\ell-1})} are related by 
\[\h{\phi_\ell(f_{\ell-1})}(\rho_i)=\Phi\brbig{\h {f_{\ell-1}}(\rho_i)}\]
for some fixed set of linear maps \m{\seqi{\Phi}}. 
Furthermore, by Lemma \ref{lem: GCNN7}, each \m{\Phi_i} must be of the form \m{M\mapsto MB_i} for 
some appropriate matrix \m{B_i\tin \CC^{d_\rho\times d_\rho}}. 
If we then define \m{\chi_\ell} as the inverse Fourier transform of \m{(\seqi{B})}, then 
by the convolution theorem (Proposition \ref{prop: convo}), \m{\phi_\ell(f_{\ell-1})=\f_{\ell-1}\ast \chi}, 
confirming that \m{\Ncal} is a G-CNN. 
The extension of this result to the vector valued case, \m{f_\ell\colon\Xcal_\ell\to V_\ell}, is straightforward. 
\end{pfof}


\ignore{
\framedef{
\begin{thm}
If \m{G} is a compact group and \m{\Ncal} is a \m{G}-CNN, 
then \m{\Ncal} is equivariant to the action of \m{G} 
in the sense of Definition \ref{def: equivariant NN}. 
\end{thm}
}
\bigskip			
}
\ig{
little bit of representation theory. 
In Section \ref{sec: fourier} we introduced representations very concretely as mappings from 
a group to some set of matrices, \m{\rho\colon G\to\CC^{d\times d}}. 
More generally, we can think of \m{\rho} as a system of linear maps 
\m{\cbrN{\rho(g)\colon U\to U}_{g\in G}} where \m{U} is some underlying vector space.  
In other words, the concept of a representation of \m{G} on \m{U} is the same  
as that of a \emph{linear action} \m{\cbrN{T_g\colon U\to U}_{\gin G}}. 
In particular, any translation action induced from an action of \m{G} on \m{\Xcal} 
(Definition \ref{def: translation}) 
is a representation of \m{G} on \m{U=\lins{\Xcal}{\Ycal}} 
(however, not every representation is necessarily a translation action).  
In the following we will go back and forth between these two alternative views of representations. 

Moving the emphasis from the maps to the space, 
if \m{\cbrN{\rho(g)\colon U\to U}_{g\in G}} is a representation of \m{G}, then 
we say that \m{U} is a \m{G}\emph{--module}, 
and we say that is irreducible if it has  
no proper subspace \m{W\<\subset U} such that \m{\rho(g)(W)\subseteq W} for all \m{g\tin G}. 
Two \m{G}--modules \m{U} and \m{V} with corresponding 
representations \m{\cbrN{\rho(g)\colon U\to U}_{g\in G}} and 
\m{\cbrN{\rho'(g)\colon V\to V}_{g\in G}} are equivalent if there is an invertible map 
\m{\psi\colon U\to V} such that \m{\psi\circ \rho(g)=\rho'(g)\circ\psi} for all \m{g\tin G}. 
We start our proof by recalling three basic results from representation theory. 
}
\ignore{
\begin{lem}\textbf{(Complete decomposability)} 
Let \m{G} be a compact group and \m{U} be a \m{G}--module. 
Then \m{U} can be reduced into a direct sum of a countable number of subspaces 
\[
U=U^1\oplus U^2\oplus \ldots, 
\]
where each \m{U^i} is an irreducible \m{G}--module. 
Now let \m{\Rcal} be a complete set of inequivalent irreducible representations of \m{G}, and 
\m{m_\rho(U)} (with \m{\rho\tin\Rcal}) be the number of irreducible representations in the above 
decomposition that are equivalent with \m{\rho}, called the multiplicity of \m{\rho} in \m{U}. 
Then for any other decomposition \m{U={U'}{}^1\oplus U'{}^2\oplus \ldots} into irreducibles,  
the multiplicity of each \m{\rho} will be the same. 
\end{lem}
}
}

\ignore{
\begin{lem}\label{lem: GCNN3} 
Let \m{\Xcal} be a homogeneous space of a compact group \m{G},  
\m{\cbrN{\Tact_g\colon U\<\to U}_{g\in G}} the translation action of \m{G} on 
\m{U\<=\lins{\Xcal}{}}, and 
\[U=U_1\oplus U_2\oplus \ldots\]
the corresponding isotypic decomposition of \m{U}. 
For any \m{f\tin U}, let \m{\cbrN{\h f(\rho_1), \h f(\rho_2),\ldots }} 
be the Fourier transform of \m{f}, as defined in Definition \ref{def: FT homo}. 
The there exists a decomposition \m{U_i=U_i^1\oplus U_i^2\oplus \ldots} of each 
isotypic into irreducible \m{G}--modules, and an orthonormal basis 
\m{\cbrN{e_{i,j,1},\ldots,e_{i,j,d_{\rho_i}}}} for each \m{U_i^j} such that 
\m{[\h f(\rho_i)]_{j,k}=\inp{f,e_{i,j,k}}} for any \m{i,j,k}. 
\end{lem}
}


\ignore{
\framedef{
\begin{thm}
Let \m{G} be a compact group and  \m{\Ncal} an \m{L\<+1} layer feed-forward network 
in which each index set \m{\seqz{\Xcal}{L}} is a homogeneous space of \m{G}. 
If each layer of \m{\Ncal} is equivarent to the action of \m{G}, 
then \m{\Ncal} is a G-CNN. 
\end{thm}
}
}

\ignore{
\begin{lem}\label{lem: GCNN4}
Let everything be as in Lemma \ref{lem: GCNN3}, 
\m{d\<=d_{\rho_i}}, and 
\m{\Theta} be the set of allowable linear maps \m{\Phi_i\colon \h f(\rho_i)\mapsto \h f'(\rho_i)}. 
Then \m{\Theta} forms a subspace of \m{\mathrm{Hom}(\CC^{d\times d},\CC^{d\times d})}, 
the space of all linear maps \m{\CC^{d\times d}\to\CC^{d\times d}}. 
\end{lem}
\noindent 
\begin{pf}
Since \m{\cbrN{\Tact}} and \m{\cbrN{\Tact'}} are linear actions, for any two 
equivariant maps \m{\phi,\phi'\colon L(\Xcal)\to L(\Xcal')}, the linear combination 
\m{\alpha \phi+\beta\phi'} is also equivariant, so 
the set of equivariant linear maps form a linear space \m{\wbar\Theta}. 
By the correspondence between the Fourier transform and the isotypic decomposition, 
any allowable \m{\Phi_i} is just the restriction of some \m{\phi\tin \wbar{\Theta}} to \m{(U_i,V_i)}. 
Therefore, the set of allowable \m{\Phi_i}'s also form a linear space.  
\end{pf}
}

\ignore{
Letting \m{U=\lins{\Xcal_{\ell-1}}{}} and \m{V=\lins{\Xcal_\ell}{}}, with \m{U=\soperi{U}{\oplus}} 
and \m{V=\soperi{V}{\oplus}} being the corresponding isotypic decompositions, 
Lemma \ref{lem: GCNN2} tells us that \m{\phi_\ell} must take each \m{U_i} to the corresponding 
\m{V_i}, i.e., the restriction of \m{\phi_\ell} to \m{U_i} is a linear map \m{\phi_\ell\vert_{U_i}\colon U_i\to V_i}. 

Now let \m{\m{\cbrN{\h f(\rho_1), \h f(\rho_2),\ldots }}} be the Fourier transform of 
\m{f\tin U} (as defined in Definition \ref{def: FT homo}) and \m{\cbrN{\h f'(\rho_1), \h f'(\rho_2),\ldots }} 
the Fourier transform of \m{f':=\phi_\ell(f)\tin V}. 
By Lemma \ref{lem: GCNN3}, \m{\h f(\rho_i)} captures exactly the part of \m{f} that falls in \m{U_i} 
(and similarly for \m{\h f'(\rho_i)}), so in ``Fourier space'', \m{\phi_\ell} must take the form 
of a sequence of linear mappings \m{\cbrN{\h \phi_\ell^i\colon \h f(\rho_i)\to \h f'(\rho_i)}_{i=1,2,\ldots}}.  
}

\section{Examples of algebraic convolution in neural networks}\label{sec: examples}

We are not aware of any prior papers that have exposed the above algebraic theory of equivariance 
and convolution in its full generality. 
However, there are a few recent publications that implicitly exploit these ideas in specific contexts. 

\subsection{Rotation equivariant networks} 

In image recognition applications it is a natural goal to achieve equivariance to both translation 
and rotation. 
The most common approach is to use CNNs, but with filters that are replicated at a certain number of 
rotational angles (typically multiples of 90 degrees), connected in such as a way as to achieve 
a generalization of equivariance called \emph{steerability}. Steerability also has a group theoretic 
interpretation, which is most lucidly explained in \citep{Cohen2017}. 
 
The recent papers \citep{Marcos2017} and \citep{Worrall2017} extend these architectures by  
considering continuous rotations at each point of the visual field. 
Thus, putting aside the steerability aspect for now and only considering the behavior of the network 
at a single point, both these papers deal with the case where \m{G=\SO(2)} (the two dimensional rotation group) 
and \m{\Xcal} is the circle \m{S^1}. 
The group \m{\SO(2)} is commutative, therefore its irreducible representations are one dimensional, and 
are, in fact, \m{\rho_j(\theta)=e^{2\pi\iota j \theta}}, where \m{\iota=\sqrt{-1}}. 
While not calling it a group Fourier transform, \citet{Worrall2017} explicitly expand the local activations 
in this basis and scale them with weights, which, by virtue of Proposition \ref{prop: convo}, 
amounts to convolution on the group, as prescribed by our main theorem.

The form of the nonlinearity in \citep{Worrall2017} is different from that prescribed in Definition 
\ref{def: equivariant NN}, which leads to a coupling between the indices of the Fourier components in any 
path from the input layer to the output layer. This is compensated by what they call their 
``equivariance condition'', asserting that only Fourier components for which \m{M=\sum_{\ell}j_\ell} is the 
same may mix. This restores equivariance in the last layer, but analyzing it group 
theoretically is beyond the scope of the present paper. 

\subsection{Spherical CNNs}

Very close in spirit to our present exposition are the recent papers 
\citep{SphericalCNN2018,SphericalCNN2018arxiv}, 
which propose convolutional architectures for recognizing images painted on the sphere, 
satisfying equivariance with respect to rotations. 
Thus, in this case, \m{G\<=\SO(3)}, the group of three dimensional rotations, and \m{\Xcal_\ell} is the 
sphere, \m{S^2}. 

The case of rotations acting on the sphere is one of the textbook examples of continuous group actions. 
In particular, letting \m{x_0} be the North pole, we see that two-dimensional rotations in the \m{x}--\m{z} 
plane fix \m{x_0}, therefore, \m{S^2} is identified with the quotient space \m{\SO(3)/\SO(2)}. 
The irreducible representations of \m{\SO(3)} are given by the so-called Wigner matrices. 
The \m{\ell}'th irreducible representation is \m{2\ell\<+1} dimensional and of the form 
\[[\rho_\ell(\theta,\phi,\psi)]_{m,m'}=e^{-\iota m'\phi}\,d^\ell_{m'\nts,m}\nts(\theta)\,e^{-\iota m\psi}, \]
where \m{m,m'\tin\cbrN{-\ell,\ldots,\ell}}, \m{(\theta,\phi,\psi)} are the Euler angles of the rotation 
and the \m{d^\ell_{m'\nts,m}(\theta)} funcion is related to the spherical harmonics. 
It is immediately clear that on restriction to \m{\SO(2)} (corresponding to \m{\theta,\phi=0}) 
only the middle column in each of these matrices reduces to the trivial representation of 
\m{\SO(2)}, therefore, by Proposition \ref{prop: ft quotient}, in the case \m{f\colon \SO(3)/\SO(2)\to\CC}, 
only the middle column of each \m{\h f(\rho_\ell)} matrix will be nonzero. 
In fact, up to constant scaling factors, the entries in that middle column  
are just the customary spherical harmonic expansion coefficients.

\citet{SphericalCNN2018} explicitly make this connection between spherical harmonics and \m{SO(3)} Fourier 
transforms, and store the activations in terms of this representation. 
Moreover, just like in the present paper, they define convolution in terms of the noncommutative 
convolution theorem (Proposition \ref{prop: convo}), use pointwise nonlinearities, and prove that 
the resulting neural network is \m{\SO(3)}--equivariant. However, they do not prove the converse, 
i.e., that equivariance implies that the network \emph{must} be convolutional. 
To apply the nonlinearity, the algorithm presented in \citep{SphericalCNN2018} requires 
repeated forward and backward \m{\SO(3)} fast Fourier transforms. 
While this leads to a non-conventional architecture, the discussion echoes our observation that 
when dealing with continuous symmetries such as rotations, one must generalize to 
more abstract ``continuous'' neural networks, as afforded by Definition \ref{def: equivariant NN}.

\subsection{Message passing neural networks}

There has been considerable interest in extending the convolutional network formalism to learning from 
graphs \citep{Niepert2016,Defferrard2016,Duvenaud2015}, and the current consensus for approaching 
this problem is to use neural networks based on the message passing idea \citep{Riley2017}. 
Let \m{\Gcal} be a graph with \m{n} vertices. 
Message passing neural networks (MPNNs) are usually presented in terms of an iterative process, where in 
each round \m{\ell}, each vertex \m{v} collects the labels of its neighbors \m{\sseq{w}{k}}, 
and updates its own label \m{\tilde f_v} according to a simple formula such as 
\begin{equation*}
\tilde f^\ell_v=\Phi\brbig{\tilde f^{\ell-1}_{w_1}+\ldots+\tilde f^{\ell-1}_{w_k}}.\vspace{-3pt}
\end{equation*}
An equivalent way of seeing this process, however, is in terms of the ``effective receptive fields'' \m{\Scal^\ell_v} 
of each vertex at round \m{\ell}, i.e., the set of all vertices from which information can propagate to 
\m{v} by round \m{\ell}. 

MPNNs can also be viewed as group convolutional networks. 
A receptive field of size \m{k} is just a subset 
\m{\cbrN{\sseq{s}{k}}\subseteq\cbrN{\oneton{n}}}, and the symmetric group \m{\Sn} 
(the group of permutations of \m{\cbrN{\oneton{n}}}) acts on the set of such subsets transitively by 
\vspace{-8pt}
\begin{equation*}
\cbrN{\sseq{s}{k}}\overset{\sigma}{\mapsto} \cbrN{\sigma(s_1),\ldots,\sigma(s_k)}\qquad \sigma\tin\Sn.
\vspace{-2pt}
\end{equation*}
Since permuting the \m{n-k} vertices \m{not} in \m{\Scal} amongst themselves, 
as well as permuting the \m{k} vertices that are in \m{\Scal} both leave \m{\Scal} invariant, 
the stablizier of this action is \m{\Sbb_{n-k}\times \Sbb_k}.  
Thus, the set of all \m{k}-subsets of vertices is identified with the quotient space 
\m{\Xcal=\Sn/(\Sbb_{k}\times \Sbb_{n-k})}, and the labeling function for \m{k}-element 
receptive fields is identified with \m{f^k\colon \Xcal\to\CC}. 
Effectively, this turns the MPNN into a generalized feed-forward network in the sense of Definition 
\ref{def: equivariant NN}. Note that \m{f^k} is a redundant representation of the labeling 
function because \m{\Sn/(\Sbb_{k}\times \Sbb_{n-k})} also includes subsets that do not correspond to 
contiguous neighborhoods. However this is not a problem because for such \m{\Scal} we simply set 
\m{f^k(\Scal)\<=0}. 

The key feature of the message passing formalism is that, by construction, it ensures that the 
\m{\smash{\tilde f^\ell_v}} labels only depend on the graph topology and are invariant to renumbering 
the vertices of \m{\Gcal}. In terms of our ``\m{k}--subset network'' this means that each \m{f^k} must 
be \m{\Sbb_n}--equivariant.
Thus, in contrast to the previous two examples, 
now each index set 
\begin{mequation}\label{eq: MPNN}
\Xcal_\ell\<=\Sn/(\Sbb_{n-\ell}\times\Sbb_\ell)
\end{mequation}
is different. The form of the corresponding convolutions 
\m{\lins{\Xcal_{\ell-1}}{V_{\ell-1}}\to \lins{\Xcal_\ell}{V_\ell}} are best described in the Fourier 
domain. Unfortunately, 
the representation theory of symmetric groups is beyond the scope of the present paper \cite{Sagan}. 
We content ourselves by stating that the irreps of \m{\Sn} are indexed by so-called 
integer partitions, \m{(\sseq{\lambda}{m})}, where \m{\lambda_1\geq\ldots\geq\lambda_m} and 
\m{\sum_i\lambda_i=n}. Moreover, the structure of the Fourier transform of a function 
\m{f\colon \Sn/(\Sbb_{n-\ell}\times\Sbb_\ell)} dictated by Proposition \ref{prop: ft quotient}  
in this case is that each of the Fourier matrices 
are zero except for a single column in each of the \m{\h f((n\<-p,p))} components, where 
\m{0 \leq p\leq \ell}. 
The main theorem of our paper dictates that the linear map \m{\phi_\ell} in each layer must be a convolution.  
In the case of Fourier matrices with such extreme sparsity structure, 
this means that each of the \m{\ell\<+1} Fourier matrices can be multiplied by a 
scalar, \m{\chi^\ell_p}. These are the learnable parameters of the network. 
A real MPNN of course has multiple channels and various corresponding parameters, which could also 
be introduced in the \m{k}--subset network. 
The above observation about the form of \m{\chi_\ell} is nonetheless interesting, because it at once implies 
that permutation equivariance is a severe constraint the significantly limits the form of the convolutional 
filters, yet the framework is still richer than traditional MPNNs where the labels of the 
neighbors are simply summed.

 
\section{Conclusions}

Convolution has emerged as one of the key organizing principles of deep neural network architectures. 
Nonetheless, depending on their background, the word ``convolution'' means different things to different 
researchers. 
The goal of this paper was to show that in the common setting when there is a group 
acting on the data that the architecture must be equivariant to, convolution has a specific 
mathematical meaning that has far reaching consequences: 
we proved that a feed forward network is equivariant to the group action  
if and only if it respects this notion of convolution. 

Our theory gives a clear prescription to practitioners on how to design 
neural networks for data with non-trivial symmetries, such as data on the sphere, etc.. 
In particular, we argue for Fourier space representations, similar to those that have 
appeared in \citep{Worrall2017,SphericalCNN2018,SphericalCNN2018arxiv}), 
and, even more recently, since the submission of the original version of the present paper in 
\citep{TensorFieldNetworks2018arxiv,Nbody2018arxiv,Weiler2018arxiv}. 

\subsection*{Acknowledgements}
\vspace{-3pt}
This work was supported in part by DARPA Young Faculty Award D16AP00112. 


\bibliography{mcnn,supp}
\bibliographystyle{icml2018}

\appendix

{\Large \textbf{Appendix}}

\section{Background from group and representation theory}\label{backgrnd}

For a more detailed background on representation theory, we point the reader to Serre, 1977.

\paragraph{Groups.} A \df{group} is a set \m{G} endowed with an operation \m{G\times G\to G} 
(usually denoted multiplicatively) obeying the following axioms: 
\begin{compactenum}[~G1.]
\item for any \m{g_1,g_2\tin G},~ \m{g_1 g_2\tin G} (closure);
\item for any \m{g_1,g_2,g_3\tin G},~ \m{g_1(g_2 g_3)=(g_1 g_2)g_3} (associativity);
\item there is a unique \m{e\tin G}, called the \df{identity} of \m{G}, 
such that \m{eg=ge=g} for any \m{u\tin G};
\item for any \m{g\tin G}, there is a corresponding element \m{g^{-1}\!\tin G} called the \df{inverse} 
of \m{g}, such that \m{g\ts g^{-1}=g^{-1}g=e}.
\end{compactenum}
We do \emph{not} require that the group operation be commutative, i.e., in general, \m{g_1 g_2\neq g_2 g_1}. 
Groups can be finite or infinite, countable or uncountable, compact or non-compact. 
While most of the results in this paper would generalize to any compact group, to keep the exposition 
as simple as possible,  
throughout we assume that \m{G} is finite or countably infinite. 
As usual, \m{\abs{G}} will denote the size (cardinality) of \m{G}, sometimes also called the \df{order} of the group. 
A subset \m{H} of \m{G} is called a \df{subgroup} of \m{G}, denoted \m{H\leq G}, if \m{H} itself forms a group 
under the same operation as \m{G}, i.e., if for any \m{g_1,g_2\tin H}, \m{g_1 g_2\tin H}. 

\paragraph{Homogeneous Spaces.}

\begin{defn}
Let \m{G} be a group acting on a set \m{\Xcal}. We say that \m{\Xcal} is a \df{homogeneous space} of 
\m{G} if for any \m{x,y\tin \Xcal}, there is a \m{g\tin G} such that \m{y\<=g(x)}. 
\end{defn}

\noindent 
The significance of homogeneous spaces for our purposes is that once we fix the ``origin'' \m{x_0}, 
the above correspondence between points in \m{\Xcal} and the group elements that map \m{x_0} to them 
allows to lift various operations on the homogeneous space to the group. 
Because expressions like \m{g(x_0)} appear so often in the following, we introduce the shorthand 
\m{[g]_{\Xcal}\!:=\!g(x_0)}. Note that this hides the dependency on the (arbitrary) choice of \m{x_0}. 

For some examples, we see that \m{\ZZ^2} is a homogeneous space of itself with respect to the trivial action 
\m{(i,j)\mapsto (g_1\<+i,g_2\<+j)}, and the sphere is a homogeneous space of the rotation group 
with respect to the action:
\begin{equation}\label{eq: rot}
x\mapsto R(x) \qquad  R(x)=R x \qquad x\tin S^2,
\end{equation}

On the other hand, the entries of the adjacency matrix 
are \emph{not} a homogeneous space of \m{\Sn} with respect to 
\begin{equation}\label{eq: A action}
(i,j)\mapsto (\sigma(i),\sigma(j))\qqquad \sigma\tin\Sn. 
\end{equation} , 
because if we take some \m{(i,j)} with \m{i\<\neq j}, 
then \ref{eq: A action} can map it to any other \m{(i',j')} with \m{i'\<\neq j'}, but not to any 
of the diagonal elements, where \m{i'\<=j'}. 
If we split the matrix into 
its ``diagonal'', and ``off-diagonal'' parts, individually these two parts are homogeneous spaces.

\paragraph{Representations.}
A (finite dimensional) \df{representation} of a group \m{G} over a field \m{\FF} 
is a matrix-valued function \m{\rho\colon G\to\FF^{d_\rho\times d_\rho}} 
such that \m{\rho(g_1)\ts\rho(g_2)=\rho(g_1 g_2)} for any \m{g_1,g_2\tin G}. 
In this paper, unless stated otherwise, we always assume that \m{\FF\<=\CC}.  
A representation \m{\rho} is said to be \df{unitary} if \m{\rho(g^{-1})=\rho(g)^\dag} 
for any \m{g\tin G}. 
One representation shared by every group is the \df{trivial representation} 
\m{\rho_{\tr}} that simply evaluates to the one dimensional matrix \m{\rho_\tr(g)=(1)} on 
every group element.

\paragraph{Equivalence, reducibility and irreps.} 
Two representations \m{\rho} and \m{\rho'} of the same dimensionality \m{d}  
are said to be \df{equivalent} if for some invertible matrix 
\m{Q\tin\CC^{d\times d}},\; \m{\rho(g)=Q^{-1}\nts\rho'(g)\,Q} for any \m{g\tin G}. 
A representation \m{\rho} is said to be \df{reducible} if  
it decomposes into a direct sum of smaller representations in the form  
\begin{align*}
\rho(g) \\
& = Q^{-1}\br{\rho_1(g)\<\oplus\rho_2(g)}\, Q \\
& = Q^{-1}\nts \br{
	\begin{array}{c|c}
	\rho_1(g)&0\\
	\hline
	0&\rho_2(g)
	\end{array}
}Q  \qquad \forall\, g \tin G
\end{align*}
for some invertible matrix \m{Q\tin\CC^{d_\rho\times d_\rho}}. 
We use \m{\Rcal_G} to denote a complete set of inequivalent irreducible representations of \m{G}. 
However, since this is quite a mouthful, in this paper we also use the alternative term 
\df{system of irreps} to refer to \m{\Rcal_G}.  
Note that the choice of irreps in \m{\Rcal_G} is far from unique, since each \m{\rho\tin \Rcal_G} can 
be replaced by an equivalent irrep \m{Q^\top\nts\nts \rho(g)\ts Q}, where \m{Q} is any orthogonal matrix 
of the appropriate size. 

\paragraph{Complete reducibility and irreps.}
Representation theory takes on its simplest form when \m{G} is compact (and \m{\FF=\CC}).  
One of the reasons for this is that it is possible to prove (``theorem of complete reducibility'') that 
any representation \m{\rho} of a compact group 
can be reduced into a direct sum of irreducible ones, i.e., 
\begin{equation}\label{eq: repr-decomp}
\rho(g)=
Q^{-1}\br{\rho_{(1)}(g)\<\oplus\rho_{(2)}(g)\oplus \ldots\oplus \rho_{(k)}(g)}\, Q, g\tin G
\end{equation}
for some sequence \m{\rho_{(1)},\rho_{(2)},\ldots,\rho_{(k)}}
of irreducible representations of \m{G} and some \m{Q\tin\CC^{d\times d}}. 
In this sense, for compact groups, \m{\Rcal_G} plays a role very similar to the primes in arithmetic. 
Fixing \m{\Rcal_G}, the number of times that a particular \m{\rho'\tin\Rcal_G} appears in \rf{eq: repr-decomp} 
is a well-defined quantity called the \df{multiplicity} of \m{\rho'} in \m{\rho}, denoted \m{m_{\rho}(\rho')}.  
Compactness also has a number of other advantages: 
\begin{compactenum}
	\item When \m{G} is compact, \m{\Rcal_G} is a countable set, therefore we can refer to the individual 
	irreps as \m{\seqi{\rho}}. (When \m{G} is finite, \m{\Rcal_G} is not only countable but finite.) 
	\item The system of irreps of a compact group is essentially unique in the sense that if \m{\Rcal_G'} is 
	any other system of irreps, then there is a bijection \m{\phi\colon \Rcal_G\to\Rcal'_G} 
	mapping each irrep \m{\rho\tin\Rcal_G} to an equivalent irrep \m{\phi(\rho)\tin\Rcal_G'}. 
	\item When \m{G} is compact, \m{\Rcal_G} can be chosen in such a way that each \m{\rho\tin\Rcal} is unitary.   
\end{compactenum}

\paragraph{Restricted representations.} 
Given any representation \m{\rho} of \m{G} and subgroup \m{H\leq G}, the \emph{restriction} of \m{\rho} to 
\m{H} is defined as the function \m{\rho\vert_H\colon H\to\CC^{d_\rho\times d_\rho}}, 
where \m{\rho\vert_H(h)=\rho(h)} for all \m{h\tin H}. 
It is trivial to check that \m{\rho\vert_H} is a representation of \m{H}, but, in general, it is not 
irreducible (even when \m{\rho} itself is irreducible). 

\paragraph{Fourier Transforms.}
In the Euclidean domain convolution and cross-correlation have close relationships 
with the Fourier transform
\begin{equation}\label{eq: Fourier classical}
\h f(k)=\int e^{-2\pi \iota kx}\,f(x)\:dx,
\end{equation} 
where \m{\iota} is the imaginary unit, \m{\sqrt{-1}}. 
In particular, the Fourier transform of \m{f\ast g} is just the pointwise product of the Fourier 
transforms of \m{f} and \m{g}, 
\begin{equation}\label{eq: convo thm}
\h{f\ast g}(k)=\h{f}(k)\,\h{g}(k),
\end{equation}
while cross-correlation is 
\begin{equation}\label{eq: corr thm}
\h{f\star g}(k)=\h f(k)^\ast\:\h{g}(k).
\end{equation}

The concept of \emph{group representations} (see Section \ref{backgrnd}) 
allows generalizing the Fourier transform to any compact group. 
The \df{Fourier transform} of \m{f\colon G\to\CC} \ig{(w.r.t. \cbrN{\seqi{\rho}}) }is defined as: 
\begin{equation}\label{eq: gp-ft}
\h f(\rho_i)=\int_G \rho_i(u)\,f(u)\:d\mu(u), \qqquad i=1,2,\ldots, 
\end{equation}
which, in the countable (or finite) case simplifies to 
\begin{equation}\label{eq: gp-ft2}
\h f(\rho_i)=\sum_{u\in G} f(u)\ts \rho(u), \qqquad i=1,2,\ldots.
\end{equation}
Despite \m{\RR} not being a compact group, \rf{eq: Fourier classical} can be seen as a 
special case of \rf{eq: gp-ft}, since \m{e^{-2\pi \iota kx}} trivially obeys 
\m{e^{-2\pi \iota k(x_1+x_2)}=e^{-2\pi \iota kx_1} e^{-2\pi \iota kx_2}}, and 
the functions \m{\rho_k(x)=e^{-2\pi \iota kx}} are, in fact, 
the irreducible representations of \m{\RR}. 
The fundamental novelty in \rf{eq: gp-ft} and \rf{eq: gp-ft2} 
compared to \rf{eq: Fourier classical}, however, is that since, in general 
(in particular, when \m{G} is not commutative),   
irreducible representations are matrix valued functions, 
each ``Fourier component'' \m{\h f(\rho)} is now a matrix. 
In other respects, Fourier transforms on groups behave very similarly to classical Fourier transforms. 
For example, we have an inverse Fourier transform 
\[f(u)=\ovr{\abs{G}}\sum_{\rho\in\Rcal} d_\rho\,\tr\sqbbig{f(\rho)\ts \rho(u)^{-1}},  
\]
and also an analog of the convolution theorem, which is stated in the main body of the paper.

\section{Convolution of vector valued functions}

Since neural nets have multiple channels, we need to further 
extend equations 6-12 to vector/matrix valued functions. 
Once again, there are multiple cases to consider. 
\\

\newcommand{\Hom}{\mathop{\textrm{Hom}}}

\begin{defn}\label{def: convo homov}
Let \m{G} be a finite or countable group, and \m{\Xcal} and \m{\Ycal} be (left or right) quotient spaces of \m{G}. 
\setlength{\plitemsep}{3pt}
\begin{compactenum}[~1.] 
\item If \m{f\colon \Xcal\to \CC^m}, and \m{g\colon \Ycal\to \CC^m}, we define \m{f\ast g\colon G\to\CC} with 
\vspace{-7pt}  
\begin{equation}\label{eq: convo homov0}
(f\ast g)(u)=\sum_{v\in G} f\tup^G(uv^{-1})\cdot g\tup^G(v), 
\vspace{-12pt}
\end{equation}
where \m{\cdot} denotes the dot product.  
\item If \m{f\colon \Xcal\to \CC^{n\times m}}, and \m{g\colon \Ycal\to \CC^m}, we define \m{f\ast g\colon G\to\CC^n} with  
\vspace{-7pt}
\begin{equation}\label{eq: convo homo1}
(f\ast g)(u)=\sum_{v\in G} f\tup^G(uv^{-1})\,\times\,  g\tup^G(v), 
\vspace{-7pt}
\end{equation}
where \m{\times} denotes the matrix/vector product.   
\item If \m{f\colon \Xcal\to \CC^{m}}, and \m{g\colon \Ycal\to \CC^{n\times m}}, we define \m{f\ast g\colon G\to\CC^m} with 
\vspace{-7pt}
\begin{equation}\label{eq: convo homo2}
(f\ast g)(u)=\sum_{v\in G}   f\tup^G(uv^{-1})\: \tilde\times\: g\tup^G(v), 
\vspace{-7pt}
\end{equation}
where \m{\V v \tilde\times A} denotes the ``reverse matrix/vector product'' \m{A\V v}.   
\end{compactenum}
Since in cases 2 and 3 the nature of the product is clear from the definition of \m{f} and \m{g}, 
we will omit the \m{\times} and \m{\tilde \times} symbols. 
The specializations of these formulae to the cases of Equations 6-12 are as to be expected.  
\end{defn}

\section{Proof of Proposition 1}

\noindent

Proposition 1 has three parts. To proceed with the proof, we introduce two simple lemmas.

Recall that if \m{H} is a subgroup of \m{G}, a function \m{f\colon G\to\CC} is called 
\df{right} \m{H}--\df{invariant} if \m{f(uh)=f(u)} for all \m{h\tin H} and all \m{u\tin G}, and it is called 
\df{left} \m{H}--\df{invariant} if \m{f(hu)=f(u)} for all \m{h\tin H} and all \m{u\tin G}. 

\begin{lem}\label{lem: homo-inv}
	Let \m{H} and \m{K} be two subgroups of a group \m{G}. Then
	\begin{compactenum}[~~1.]
		\item If \m{f\colon G/H\to\CC}, then \m{f\tup^G\colon G\to\CC} is right \m{H}--invariant. 
		\item If \m{f\colon H\backslash G\to\CC}, then \m{f\tup^G\colon G\to\CC} is left \m{H}--invariant. 
		\item If \m{f\colon K\backslash G/H\to\CC}, then \m{f\tup^G\colon G\to\CC} is right \m{H} invariant and 
		left \m{K}--invariant. 
	\end{compactenum}
\end{lem}

\begin{lem}\label{lem: ft restr}
	Let \m{\rho} be an irreducible representation of a countable group \m{G}. 
	Then \m{\sum_{u\in G}\rho(u)=0} unless \m{\rho} is the trivial  
	representation, \m{\rho_{\tr}(u)=(1)}. 
\end{lem}

\noindent
\begin{pf}
	Let us define the functions 
	\m{r^\rho_{i,j}(u)=[\rho(u)]_{i,j}}. 
	Recall that for \m{f,g\colon G\to\CC}, the inner product \m{\inp{f,g}} is defined 
	\m{\inp{f,g}=\sum_{u\in G}f(u)^\ast g(u)}. 
	The Fourier transform of a function \m{f} can then be written element-wise as 
	\m{[\h f(\rho)]_{i,j}=\inpN{{r^{\rho}_{i,j}}^\ast,f}}.
	However, since the Fourier transform is a unitary transformation, 
	for any \m{\rho,\rho'\tin\Rcal_G}, unless \m{\rho\<=\rho'}, \m{i\<=i'} and \m{j\<=j'}, 
	we must have \m{\inpN{r^{\rho}_{i,j},r^{\rho'}_{i',j'}}=0}. 
	In particular, 
	\begin{mequation*}
		\sqbbig{\sum_{u\in G}\rho(u)}_{i,j}=\inpN{r^{\rho_{\tr}}_{1,1},r^{\rho}_{i,j}}=0,
	\end{mequation*}
	unless \m{\rho=\rho_{\tr}} (and \m{i\<=j\<=1}). 
\end{pf}

\noindent 
Now recall that given an irrep \m{\rho} of \m{G}, the \emph{restriction} of \m{\rho} to \m{H} is 
\m{\rho\vert_H\colon H\to\CC^{d_\rho\times d_\rho}}, where \m{\rho\vert_H(h)=\rho(h)} for all \m{h\tin H}. 
It is trivial to check that \m{\rho\vert_H} is a representation of \m{H}, but, in general, it is not 
irreducible. Thus, by the Theorem of Complete Decomposability (see section \ref{backgrnd}), 
it must decompose in the form \m{\rho\vert_H(h)=Q(\mu_1(h)\oplus \mu_2(h)\oplus \ldots \oplus \mu_k(h))Q^\dag} 
for some sequence \sseq{\mu}{k} of irreps of \m{H} and some unitary martrix \m{Q}. 
In the special case when the irreps of 
\m{G} and \m{H} are adapted to \m{H\leq G}, however, \m{Q} is just the unity.

This is essentially the case that we consider in Proposition 1. Now, armed with the above lemmas, we are in a position 
to prove Proposition 1.

\subsubsection{Proof of Part 1}\label{pf: Part1Prop1}

\noindent 
\begin{pf}
	The fact that any \m{u\tin G} can be written uniquely as \m{u\<=gh} where \m{g} is the representative of 
	one of the \m{gH} cosets and \m{h\tin H} immediately tells us that \m{\h f(\rho)} factors as
	\vspace{-5pt} 
	\begin{align*}\label{eq: ft-homo1-fact}
	\h f(\rho)
	& =\sum_{u\in G} f\tup^{G}(u)\ts \rho(u) = \sum_{x\in G/H} \sum_{h\in H} f\tup^{G}(\crepr{x}h)\ts \rho(\crepr{x}h) \\
	& = \sum_{x\in G/H} \sum_{h\in H} f(x)\ts \rho(\crepr{x}h) = \sum_{x\in G/H} \sum_{h\in H} f(x)\ts \rho(\crepr{x})\rho(h)\\
	& = \sum_{x\in G/H} f(x)\ts \rho(\crepr{x}) \sqbBig{\sum_{h\in H}\rho(h)}.
	\end{align*}
	However, \m{\rho(h)=\mu_1(h)\oplus \mu_2(h)\oplus \ldots \oplus \mu_k(h)} for some sequence 
	of irreps \m{\sseq{\mu}{k}} of \m{H}, so 
	\[\sum_{h\in H}\rho(h)=
	\sqbBig{\sum_{h\in H}\mu_1(h)}\oplus 
	\sqbBig{\sum_{h\in H}\mu_2(h)}\oplus \ldots \oplus  
	\sqbBig{\sum_{h\in H}\mu_k(h)}, 
	\]
	and by Lemma \ref{lem: ft restr} each of the terms in this sum where \m{\mu_i} is \emph{not} the 
	trivial representation (on \m{H}) is a zero matrix, zeroing out all the corresponding 
	columns in \m{\h f(\rho)}.  
\end{pf}

\subsubsection{Proof of Part 2}

\noindent 
\begin{pf}
	Analogous to the proof of part 1, using \m{u=hg} and a 
	factorization similar to that of \m{\h f(\rho)} in \ref{pf: Part1Prop1} except that \m{\sum_{h\in H}\rho(h)} 
	will now multiply \m{\sum_{x\in H\backslash G} f(x)\ts \rho(\crepr{x})} from the left.  
\end{pf}

\subsubsection{Proof of Part 3}

\noindent 
\begin{pf}
	Immediate from combining case 3 of Lemma \ref{lem: homo-inv} with Parts 1 and 2 of Proposition 1. 
\end{pf}

\section{Proof of Proposition 2}

\noindent

\begin{pf}
	Let us assume that \m{G} is countable. Then 
	\begin{align*}
	\h{f\<\ast g}(\rho_i)
	& = \sum_{u\in G} \sqbBig{\sum_{v\in G} f(u v^{-1})\,g(v)}\rho_i(u) \\
	& = \sum_{u\in G} \sum_{v\in G} f(u v^{-1})\,g(v) \rho_i(uv^{-1})\rho_i(v) \\
	& = \sum_{v\in G} \sum_{u\in G} f(u v^{-1})\,g(v) \rho_i(uv^{-1})\rho_i(v) \\
	& = \sum_{v\in G} \sqbBig{\sum_{u\in G} f(u v^{-1})\,\rho_i(uv^{-1})}g(v) \rho_i(v) \\
	& = \sum_{v\in G} \sqbBig{\sum_{w\in G} f(w)\,\rho_i(w)}g(v) \rho_i(v)  \\
	& = \sqbBig{\sum_{w\in G} f(w)\,\rho_i(w)} \sqbBig{\sum_{v\in G} g(v) \rho_i(v)} \\
	& = \h f(\rho_i)\, \h g(\rho_i).
	\end{align*}
	The continuous case is proved similarly but with integrals with respect Haar measure 
	instead of sums. 
\end{pf}

\section{Proof of Theorem 1}

\subsection{Reverse Direction}
\noindent 
Proving the ``only if'' part of Theorem 1 requires concepts from representation 
theory and the notion of generalized Fourier transforms (Section \ref{backgrnd})). 
We also need two versions of Schur's Lemma. 

\begin{lem}\textbf{(Schur's lemma I)}
Let \m{\cbrN{\rho(g)\colon U\<\to U}_{g\in G}} and \m{\cbrN{\rho'(g)\colon V\<\to V}_{g\in G}} 
be two irreducible representations of a compact group \m{G}. 
Let \m{\phi\colon U\to V} be a linear (not necessarily invertible) mapping that is equivariant with these 
representations in the sense that \m{\phi(\rho(g)(u))=\rho'(g)(\phi(u))} for any \m{u\tin U}. 
Then, unless \m{\phi} is the zero map, \m{\rho} and \m{\rho'} are equivalent representations. 
\end{lem}

\begin{lem}\textbf{(Schur's lemma II)} 
Let \m{\cbrN{\rho(g)\colon U\to U}_{g\in G}} be an irreducible representation of a compact group \m{G} 
on a space \m{U}, and \m{\phi\colon U\to U} a linear map that commutes with each \m{\rho(g)} 
(i.e., \m{\rho(g)\circ \phi=\phi\circ \rho(g)} for any \m{g\tin G}). 
Then \m{\phi} is a multiple of the identity.
\end{lem}

\noindent 
We build up the proof through a sequence of lemmas. 

\begin{lem}\label{lem: GCNN1}
Let \m{U} and \m{V} be two vector spaces on which a compact group \m{G} acts 
by the linear actions \m{\cbrN{T_g\colon U\!\<\to U}_{g\in G}} and \m{\cbrN{T'_g\colon V\!\<\to V}_{g\in G}}, 
respectively. Let \m{\phi\colon U\!\<\to V} be a linear map that is equivariant with the \m{\cbrN{T_g}} 
and \m{\cbrN{T'_g}} actions, and \m{W} be an irreducible subspace of \m{U} (with respect to \m{\cbrN{T_g}}). 
Then \m{Z\<=\phi(W)} is an irreducible subspace of \m{V}, and the restriction of \m{\cbrN{T_g}} to 
\m{W}, as a representation, is equivalent with the restriction of \m{\cbrN{T'_g}} to \m{Z}. 
\end{lem}

\noindent
\begin{pf}
Assume for contradiction that \m{Z} is reducible, i.e., that it has a proper subspace \m{\Zcal\subset Z} 
that is fixed by \m{\cbrN{T'_g}} (in other words, \m{T'_g(v)\tin \Zcal} for all \m{v\tin \Zcal} and \m{g\tin G}). 
Let \m{v} be any nonzero vector in \m{\Zcal}, \m{u\tin U} be such that \m{\phi(u)\<=v},  
and \m{\mathcal{W}=\Tspan\setof{T_g(u)}{g\tin G}}. 
Since \m{W} is irreducible, \m{\mathcal{W}} cannot be a proper subspace of \m{W}, so \m{\mathcal{W}=W}. 
Thus, 
\begin{multline}
Z=\phi(\Tspan\setof{T_g(u)}{g\tin G}) \\
=\Tspan\setofN{T'_g(\phi(u))}{g\tin G}=\Tspan\setofN{T'_g(v)}{g\tin G}\subseteq\Zcal,
\end{multline}
contradicting our assumption. 
Thus, the restriction \m{\cbrN{T_g\vert_W}} of \m{\cbrN{T_g}} to \m{W} and the restriction 
\m{\cbrN{T'_g\vert_Z}} of \m{\cbrN{T'_g}} to \m{Z} are both irreducible representations, 
and \m{\phi\colon W\to Z} is a linear map that is equivariant with them.  
By Schur's lemma it follows that \m{\cbrN{T_g\vert_W}}  and \m{\cbrN{T'_g\vert_Z}} are 
equivalent representations. 
\end{pf}

\begin{lem}\label{lem: GCNN2}
Let \m{U} and \m{V} be two vector spaces on which a compact group \m{G} acts 
by the linear actions \m{\cbrN{T_g\colon U\<\to U}_{g\in G}} and \m{\cbrN{T'_g\colon V\<\to V}_{g\in G}},    
and let \m{U=\soperi{U}{\oplus}} and \m{V=\soperi{V}{\oplus}}  
be the corresponding isotypic decompositions. 
Let \m{\phi\colon U\to V} be a linear map that is equivariant with the \m{\cbrN{T_g}} and \m{\cbrN{T'_g}} 
actions. Then \m{\phi(U_i)\subseteq V_i} for any \m{i}. 
\end{lem}

\noindent 
\begin{pf}
Let \m{U_i=U_i^1\oplus U_i^2\oplus\ldots } be the decomposition of \m{U_i} into irreducible 
\m{G}--modules, and \m{V_i^j=\phi(U_i^j)}. 
By Lemma \ref{lem: GCNN1}, each \m{V_i^j} is an irreducible \m{G}--module that is equivalent with \m{U_i^j}, 
hence \m{V_i^j\subseteq V_i}. Consequently, \m{\phi(U_i)=\phi(U_i^1\oplus U_i^2\oplus\ldots)\subseteq V_i}.  
\end{pf}

\begin{lem}\label{lem: GCNN3}
Let \m{\Xcal=G/H} and \m{\Xcal'=G/K} be two homogeneous spaces of a compact group \m{G}, 
let \m{\cbrN{\Tact_g\colon L(\Xcal)\<\to L(\Xcal)}_{g\in G}} and 
\m{\cbrN{\Tact'_g\colon L(\Xcal')\<\to L(\Xcal')}_{g\in G}} be 
the corresponding translation actions, and let \m{\phi\colon L(\Xcal)\to L(\Xcal')} be a linear map that 
is equivariant with these actions.  
Given \m{f\tin L(\Xcal)} let \m{\h f} denote its Fourier transform with respect to 
a specific choice of origin \m{x_0\tin \Xcal} and system or irreps \m{\Rcal_G=\cbrN{\seqi{\rho}}}. 
Similarly, \m{\h f'} is the Fourier transform of \m{f'\tin L(\Xcal')}, 
with respect to some \m{x_0'\tin\Xcal'} and the same system of irreps. 

Now if \m{f'=\phi(f)}, then each Fourier component of \m{f'} is a linear function of the corresponding 
Fourier component of \m{f}, i.e., there is a sequence of linear maps \m{\cbrN{\Phi_i}} such that 
\begin{mequation*}
\h{f'}(\rho_i)=\Phi_i(\h f(\rho_i)).
\end{mequation*}
\end{lem}
\noindent
\begin{pf} 
Let \m{\soperi{U}{\oplus}} and \m{\soperi{V}{\oplus}} be the isotypic decompositions of \m{L(\Xcal)} 
and \m{L(\Xcal')} with respect to the \m{\cbrN{\Tact_g}} and \m{\cbrN{\Tact'_g}} actions. 
By our discussion in Section \ref{sec: isotypics}, each Fourier component \m{\h f(\rho_i)} 
captures the part of \m{f} falling in the corresponding isotypic subspace \m{U_i}. 
Similarly, \m{\h f'(\rho_j)} captures the part of \m{f'} falling in \m{V_j}.    
Lemma \ref{lem: GCNN2} tells us that because \m{\phi} is equivariant with the translation actions, 
it maps each \m{U_i} to the corresponding isotypic \m{V_i}. 
Therefore, \m{\h f'(\rho_i)=\Phi_i(\h f(\rho_i))} for some function \m{\Phi_i}. 
By the linearity of \m{\phi}, each \m{\Phi_i} must be linear. 
\end{pf}
\noindent 
Lemma \ref{lem: GCNN3} is a big step towards describing what form equivariant 
mappings take in Fourier space,
but it doesn't yet fully pin down the individual \m{\Phi_i} maps. 
We now focus on a single pair of isotypics \m{(U_i,V_i)} and the corresponding 
map \m{\Phi_i} \ig{\colon \CC^{d\times d}\to \CC^{d\times d}}taking \m{\h f(\rho_i)\mapsto \h f'(\rho_i)}. 
We will say that \m{\Phi_i} is an \emph{allowable} map if there is some equivariant \m{\phi} 
such that \m{\h{\phi(f)}(\rho_i)=\Phi_i(\h f(\rho_i))}.  
Clearly, \ig{by the linearity of equivariance, }if \m{\seqi{\Phi}} are individually allowable, then 
they are also jointly allowable. 
\\

\begin{lem}\label{lem: GCNN5}
All linear maps of the form \m{\Phi_i\colon M\mapsto MB} where \m{B\tin\CC^{\delta\times \delta}} are allowable. 
\end{lem}
\noindent 
\begin{pf}
Recall that the \m{\cbrN{\Tact_g}} action takes \m{f\<\mapsto f^g}, where \m{f^g(x)\<=f(g^{-1}x)}. 
In Fourier space, 
\vspace{-5pt}
\begin{multline}\label{eq: GCNNeq1}
\h{f^g}(\rho_i)=
\sum_{u\in G}\rho_i(u)\ts f^g\tup^G\nts(u) \\
= \sum_{u\in G}\rho_i(u)\ts f\tup^G\nts(g^{-1} u) \\
= \sum_{w\in G}\rho_i(gw)\ts f\tup^G\nts(w) \\
= \rho_i(g) \sum_{w\in G}\rho_i(w)\ts f\tup^G\nts(w) \\
= \rho_i(g)\ts \h f(\rho_i).
\vspace{-5pt}
\end{multline}
(This is actually a general result called the (left) translation theorem.) 
Thus, 
\[
\Phi_i\brbig{\h{\Tact_g(f)}(\rho_i)}=
\Phi_i\brbig{\rho_i(g)\h f(\rho_i)}=
\rho_i(g)\ts \h f(\rho_i)\ts B.
\]
Similarly, the \m{\cbrN{\Tact'_g}} action maps 
\m{\h f'(\rho_i)\mapsto g(\rho_i) \h f'(\rho_i)}, so  
\[
\Tact'_g\brbig{\Phi_i(\h f(\rho_i))}=
\Tact'_g\brbig{\h f(\rho_i)\ts B}=
\rho_i(g)\ts \h f(\rho_i)\ts B.
\]
Therefore, \m{\Phi_i} is equivariant with the \m{\cbrN{\Tact}} 
and \m{\cbrN{\Tact'}} actions. 
\end{pf}

\begin{lem}\label{lem: GCNN6}
Let \m{\Phi_i\colon M\mapsto BM} for some \m{B\tin\CC^{\delta\times \delta}}. 
Then \m{\Phi_i} is not allowable unless \m{B} is a multiple of the identity.  
Moreover, this theorem also hold in the columnwise sense that if \m{\Phi_i\colon M\to M'} such 
that \m{[M']_{\ast,j}=B_j\, [M]_{\ast,j}} for some sequence of matrices \m{\sseq{B}{d}}, 
then \m{\Phi_i} is not allowable unless each \m{B_j} is a multiple of the identity. 
\end{lem}
\noindent 
\begin{pf}
Following the same steps as in the proof of Lemma \ref{lem: GCNN5}, we now have 
\begin{multline*}
\Phi_i\brbig{\h{\Tact_g(f)}(\rho_i)}=B\ts \rho_i(g)\ts \h f(\rho_i),\\ 
\Tact'_g\brbig{\Phi_i(\h f(\rho_i))}=\rho_i(g)\ts B\ts \h f(\rho_i).
\end{multline*} 
However, by the second form of Schur's Lemma, we cannot have \m{B\ts \rho_i(g)=\rho_i(g) B} 
for all \m{g\tin G}, unless \m{B} is a multiple of the identity. 
\end{pf}

\newcommand{\alpc}[4]{\alpha_{#1,#2,#3,#4}}
\newcommand{\alpcd}[4]{\alpha'{}_{\hspace{-3pt}#1,#2,#3,#4}}
\newcommand{\alpcdd}[4]{\alpha''{}_{\hspace{-5pt}#1,#2,#3,#4}}
\newcommand{\alpcz}[4]{\alpha^{0}{}_{\hspace{-5pt}#1,#2,#3,#4}}
\begin{lem}\label{lem: GCNN7}
\m{\Phi_i} is allowable if and only if it is of the form \m{M\mapsto MB} for some 
\m{B\tin\CC^{\delta\times \delta}}.
\end{lem}
\noindent 
\begin{pf}
For the ``if'' part of this lemma,  see Lemma \ref{lem: GCNN5}.  
For the ``only if'' part, note that the set of allowable \m{\Phi_i} form a subspace of all linear maps 
\m{\CC^{\delta\times \delta}\to\CC^{\delta\times \delta}}, and any allowable \m{\Phi_i} can be expressed in the form
\[[\Phi_i(M)]_{a,b}=\sum_{c,d} \alpc{a}{b}{c}{d} M_{c,d}.\]  
By Lemma \ref{lem: GCNN6}, if \m{a\neq c} but \m{b\<=d}, then \m{\alpc{a}{b}{c}{d}\<=0}. 
On the other hand, by Lemma \ref{lem: GCNN5} if \m{a\<=c}, then \m{\alpc{a}{b}{c}{d}} 
can take on any value, regardless of the values of \m{b} and \m{d}, as long as 
\m{\alpc{a}{b}{a}{d}} is constant across varying \m{a}.

Now consider the remaining case \m{a\neq c} and \m{b\neq d}, and assume that   
\m{\alpc{a}{b}{c}{d}\neq 0} while \m{\Phi_i} is still allowable. 
Then, by Lemma \ref{lem: GCNN5}, it is possible to construct a second allowable map \m{\Phi_i'} 
(namely one in which \m{\alpcd{a}{d}{a}{b}=1} and 
\m{\alpcd{a}{d}{x}{y}=0} for all \m{(x,y)\neq(c,d)}) such that in the composite map 
\m{\Phi_i''=\Phi_i'\circ\Phi_i}, \m{\alpcdd{a}{d}{c}{d}\neq 0}. 
Thus, \m{\Phi_i''} is not allowable. However, the composition of one allowable map 
with another allowable map is allowable, contradicting our assumption that \m{\Phi_i} is allowable. 

Thus, we have established that if \m{\Phi_i} is allowable, then \m{\alpc{a}{b}{c}{d}\<=0}, unless \m{a\<=c}. 
To show that any allowable \m{\Phi_i} of the form \m{M\mapsto MB}, 
it remains to prove that additionally \m{\alpc{a}{b}{a}{d}} is constant across \m{a}. 
Assume for contradiction that \m{\Phi_i} is allowable, but for some \m{(a,e,b,d)} indices 
\m{\alpc{a}{b}{a}{d}\neq\alpc{e}{b}{e}{d}}. 
Now let \m{\Phi_0} be the allowable map that zeros out every column except column \m{d} 
(i.e., \m{\alpcz{x}{d}{x}{d}=1} for all \m{x}, but all other coefficients are zero), 
and let \m{\Phi'} be the allowable map that moves column \m{b} to column \m{d} 
(i.e., \m{\alpcd{x}{d}{x}{b}=1} for any \m{x}, but all other coeffcients are zero). 
Since the composition of allowable maps is allowable, we expect \m{\Phi''=\Phi'\circ\Phi\circ \Phi^0} 
to be allowable. 
However \m{\Phi''} is a map that falls under the purview of Lemma \ref{lem: GCNN6}, yet 
\m{\alpcdd{a}{d}{a}{d}\neq\alpcdd{e}{d}{e}{d}} (i.e., \m{M_j} is not a multiple of the identity) 
creating a contradiction. 
\end{pf}

\bigskip 
\noindent 
\begin{pfof}{Theorem 1 (reverse direction)} 
For simplicty we first prove the theorem assuming \m{\Ycal_\ell\<=\CC} for each \m{\ell}. 

Since \m{\Ncal} is a G-CNN, each of the mappings  
\m{(\actfn_\ell\circ \phi_\ell)\colon \lins{\Xcal_{\ell-1}}{}\to \lins{\Xcal_{\ell}}{}} 
is equivariant with the corresponding 
translation actions \m{\cbrN{\Tact^{\ell-1}_g}_{g\in G}} and \m{\cbrN{\Tact_g^\ell}_{g\in G}}. 
Since \m{\actfn_\ell} is a pointwise operator, this is equivalent to asserting that 
\m{\phi_\ell} is equivariant with \m{\cbrN{\Tact^{\ell-1}_g}_{g\in G}} and \m{\cbrN{\Tact_g^\ell}_{g\in G}}. 

Letting \m{\Xcal=\Xcal_{\ell-1}} and \m{\Xcal'=\Xcal_\ell}, Lemma \ref{lem: GCNN5} then tells us 
the the Fourier transforms of \m{f_{\ell-1}} and \m{\phi_\ell(f_{\ell-1})} are related by 
\[\h{\phi_\ell(f_{\ell-1})}(\rho_i)=\Phi\brbig{\h {f_{\ell-1}}(\rho_i)}\]
for some fixed set of linear maps \m{\seqi{\Phi}}. 
Furthermore, by Lemma \ref{lem: GCNN7}, each \m{\Phi_i} must be of the form \m{M\mapsto MB_i} for 
some appropriate matrix \m{B_i\tin \CC^{d_\rho\times d_\rho}}. 
If we then define \m{\chi_\ell} as the inverse Fourier transform of \m{(\seqi{B})}, then 
by the convolution theorem (Proposition 2), \m{\phi_\ell(f_{\ell-1})=f_{\ell-1}\ast \chi}, 
confirming that \m{\Ncal} is a G-CNN. 
The extension of this result to the vector valued case, \m{f_\ell\colon\Xcal_\ell\to V_\ell}, is straightforward. 
\end{pfof}

\end{document}